\pdfoutput=1

\documentclass[11pt]{article}

\usepackage[preprint]{acl}

\usepackage{times}
\usepackage{latexsym}
\usepackage{amsmath}
\usepackage{amssymb}

\usepackage[T1]{fontenc}

\usepackage[utf8]{inputenc}
\usepackage{hyperref}
\usepackage{microtype}

\usepackage{inconsolata}

\usepackage{graphicx}

\usepackage{hyperref}
\usepackage{booktabs}
\usepackage{multirow}
\usepackage{subfigure}

\title{When Meaning Stays the Same, but Models Drift: Evaluating Quality of Service under Token-Level Behavioral Instability in LLMs}

\author{
  \textbf{Xiao Li\textsuperscript{1}},
  \textbf{Joel Kreuzwieser\textsuperscript{2}},
  \textbf{Alan Peters\textsuperscript{1}}
\\
  \textsuperscript{1}Electrical and Computer Engineering, Vanderbilt University\\
  \textsuperscript{2} Radiology and Radiological Sciences, Vanderbilt University
\\
  \small{
    \textbf{Correspondence:}
    \texttt{\{xiao.li, joel.p.kreuzwieser, alan.peters\}@vanderbilt.edu}
  }
}

\begin{document}
\maketitle
\begin{abstract}
We investigate how large language models respond to prompts that differ in their token-level realization but preserve the same semantic intent, a phenomenon we call prompt variance. We propose Prompt-Based Semantic Shift (PBSS), a diagnostic framework for measuring behavioral drift in LLMs under semantically equivalent prompt rewordings. Applied to ten constrained tasks, PBSS reveals consistent, model-specific response shifts—suggesting statistical regularities linked to tokenization and decoding. These results highlight an overlooked dimension of model evaluation stability under rephrasing, and suggest that tokenization strategies and decoding dynamics may contribute to post-training quality-of-service (QoS) instability.  All code, prompts, and outputs are available at: \href{https://github.com/Xiao-Vandy/LLM-Prompt-Variance-Diagnostic-Analysis.git}{our github link}.
\end{abstract}

\section{Introduction: Prompt Equivalence, Behavioral Drift}

Large language models (LLMs) are increasingly deployed in real-world settings—from legal drafting and customer service to clinical decision support. As these systems enter high-stakes domains, concerns about consistency and controllability grow. A critical yet underexplored issue is their instability under minor prompt variations—subtle changes in phrasing that can yield divergent outputs, with consequences for reliability and user trust.

Consider a hypothetical scenario where a commercial LLM assists in clinical workflows. When prompted with:

\begin{quote}
\textit{"Give a brief explanation of this result from an MRI scan for a junior physician."}
\end{quote}

and then rephrased as:

\begin{quote}
\textit{"Can you walk me through this scan as if I were a medical trainee?"}
\end{quote}

\begin{figure}[!t]
    \centering
    \begin{minipage}[t]{0.44\linewidth}
        \centering
        \includegraphics[width=\linewidth]{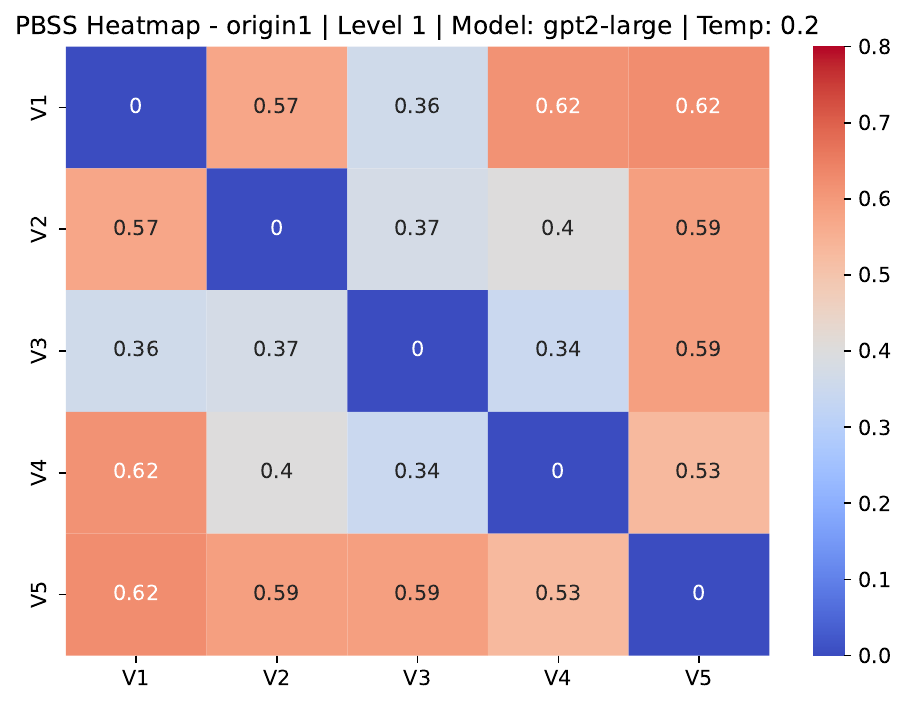}
    \end{minipage}
    \hfill
    \begin{minipage}[t]{0.44\linewidth}
        \centering
        \includegraphics[width=\linewidth]{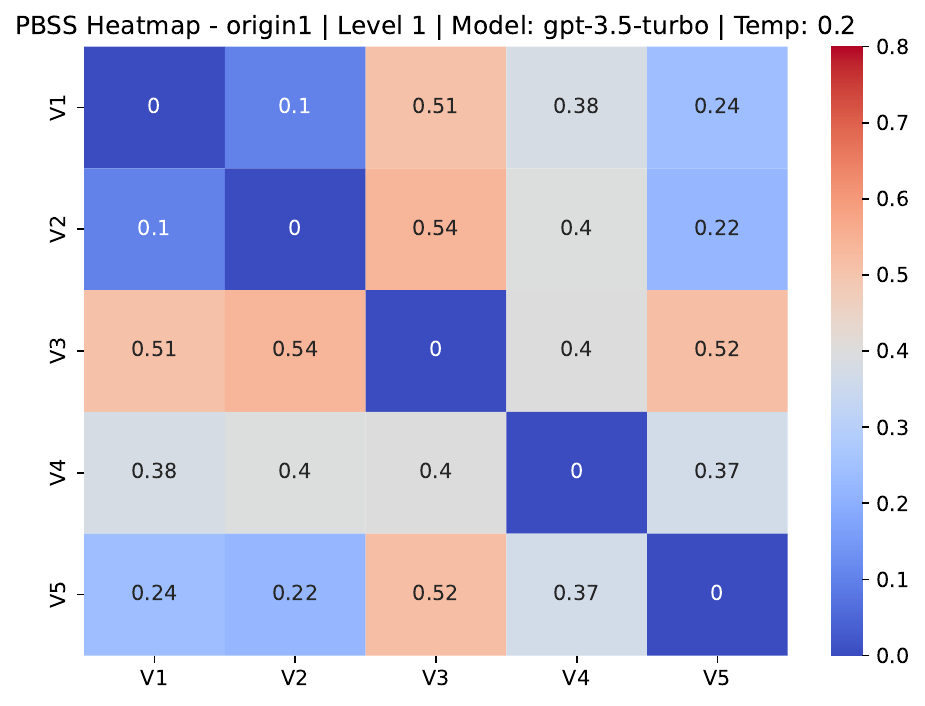}
    \end{minipage}
    \caption{PBSS heatmap showing output divergence across five semantically equivalent prompts. Red = less semantic similarity. See Section~\ref{sec:experiments}.
}
\end{figure}

The same model may hedge, speculate, or add cautionary notes absent from the original. Although the task remains constant, subtle shifts in tone, emphasis, and even factual framing emerge. These inconsistencies raise concerns not only for trust and alignment, but for safety—especially in domains like healthcare, law, and code generation, where minor rewordings can yield materially different outputs.

When applied to high-stakes domains, particularly clinical or legal—such behavioral drift is not a harmless artifact of generation. It becomes a potential failure mode. A reworded prompt that shifts diagnostic tone or factual framing may misalign the model’s output in ways that carry real-world consequences. Our framework aims to surface this instability before it turns dangerous.

It raises a broader question: \textit{How sensitive are language models to the token form, rather than the content, of their instructions?} While often dismissed as stylistic noise, we find these variations follow systematic, measurable patterns—revealing how LLMs internalize surface-level cues in unexpected ways.

\textbf{This work makes three contributions:}
\begin{itemize}
\item We formalize \textbf{Prompt-Based Semantic Shift (PBSS)} as a behavioral diagnostic framework for evaluating model sensitivity to surface-level variation in instructions—often grounded in token-level realizations. PBSS reveals significant differences across model families, validated by Kruskal-Wallis tests using three independent semantic encoders, with the most pronounced effects observed when encoders are combined.
\item We conduct a systematic analysis of prompt variance as a structured behavioral phenomenon—an angle rarely addressed in prior work, which often treats output variability as incidental or adversarial.
\item We will release a lightweight toolkit for measuring paraphrastic drift—offering scalable diagnostics across models, temperatures, and token-level prompt variants. To our knowledge, this is the first use of semantic embedding models as behavioral diagnostics for LLM stability under meaning-preserving rewordings. This work not only identifies an interpretable axis of model sensitivity, but also aims to encourage the community to more systematically investigate the underlying behavioral structure of language models.
\end{itemize}

\section{Related Work}
Recent studies have examined language model behavior from multiple angles. Our work contributes to this growing literature by focusing on semantic drift under paraphrastic variation, a phenomenon that connects behavioral instability with token-level differences in input realization. We review related work across several threads that inform this framing.
\subsection{Prompt Robustness \& Sensitivity Analysis}
Subtle changes in prompt phrasing can lead to significant shifts in LLM behavior~\cite{b1,b2,b4}, even when model outputs appear accurate. Some studies frame prompts as latent capability triggers~\cite{b1}, while others question whether LLMs meaningfully interpret instruction semantics~\cite{b2}. Alignment via fine-tuning may increase susceptibility to injection attacks~\cite{b3}, and larger models, though more capable, often become less predictable~\cite{b4}. Prompting has also been shown to exacerbate spurious correlations under distributional shift~\cite{b5}.

These findings indicate that prompt form—not just intent—can induce structured and model-specific behavioral variance. Our work builds on this insight, proposing PBSS as a diagnostic framework that quantifies semantic drift from surface-level paraphrasing.
\subsection{Instruction Tuning \& Model Behavior Shaping}

Instruction tuning—fine-tuning models on curated prompts with human supervision—has been shown to improve alignment, truthfulness, and refusal behavior across diverse tasks~\cite{b6,b7,b8,b9}. Techniques range from function-call synthesis and refusal modeling to adversarial augmentation for noisy contexts.

While tuning aims to shape model behavior via retraining, our PBSS framework provides a post-hoc diagnostic: instead of suppressing instability, it reveals structured sensitivity to semantically equivalent prompts—without modifying the model.
\subsection{Hallucination, Factual Drift \& Output Consistency}
Factual inconsistency—often termed hallucination—has been widely studied in summarization, dialogue, and QA~\cite{b10,b11}. Prior work proposes evaluation metrics aligned with human judgments~\cite{b10}, surveys domain-specific patterns~\cite{b11}, and develops detection strategies ranging from inter-sample agreement to attribution-based decoding~\cite{b12,b13}.

We position PBSS as an upstream diagnostic: if a model fails to maintain consistency under semantically equivalent prompts, its ability to preserve factual content downstream may be compromised.
\subsection{Alignment, Persona, and Jailbreak Detection}
Behavioral alignment has become a core concern in high-stakes LLM deployment, with research highlighting risks posed by prompt phrasing, persona shifts, and jailbreak attacks~\cite{b14,b15,b16,b17,b18}. Recent work documents adversarial prompt strategies, multi-turn jailbreaks, and emergent behaviors that challenge alignment even under RLHF.

Our work adopts a complementary perspective: rather than probing for failure via adversarial triggers, PBSS quantifies behavioral drift under benign, semantically equivalent prompts—offering a neutral diagnostic lens on soft-alignment instability.
\subsection{Behavioral Evaluation \& Emergent Phenomena in LLMs}
As LLMs scale, unpredictable behaviors increasingly surface—complicating evaluation and sparking interest in diagnostics beyond accuracy~\cite{b19,b20,b21}. Chain-of-thought prompting and emergent reasoning offer benchmarks for capability gains, but often mask underlying instability in structure and style.

PBSS complements this line of work by tracing behavioral drift at micro-scale: instead of highlighting new capabilities, it surfaces early signs of volatility in semantically stable contexts—offering a lightweight lens into emergent unpredictability.
\subsection{Semantic Embeddings}
Semantic embedding models such as SBERT and SimCSE are widely used for measuring textual similarity, typically in retrieval or evaluation settings~\cite{b22,b23}. While prior work compares outputs to external references, we repurpose these models as internal diagnostics—quantifying intra-model variation across semantically equivalent prompts.

PBSS reframes embedding models as stability lenses: rather than scoring correctness, we track distributional drift in how token-level representations aggregate into meaning. This enables lightweight, black-box analysis of stylistic inconsistency and latent prompt sensitivity.

\section{PBSS: A Diagnostic Framework for Token-Level Behavioral Drift}
\subsection{Why Prompt Variance Requires Token-Aware Evaluation}

We define \textbf{prompt variance} as behavioral shifts in LLM outputs triggered by surface-level paraphrases that preserve intent. These shifts are not errors or hallucinations, but drift induced by minor lexical or syntactic changes.

Critically, such variance often arises from different token sequences—variations that standard accuracy metrics ignore. We argue that prompt variance reflects a distinct behavioral axis, requiring evaluation methods sensitive to tokenization effects and rhetorical stability under paraphrasing.

\subsection{PBSS: Measuring Behavioral Sensitivity under Prompt Variation}

We define \textbf{Prompt-Based Semantic Shift (PBSS)} as a lightweight framework that measures how LLMs respond to paraphrased prompts—surface-level token changes that preserve semantic intent. PBSS treats such variation as a structured input perturbation, probing whether minor lexical or syntactic edits yield disproportionate shifts in output.

Using sentence embeddings, we compute pairwise distances between model responses to these paraphrases. The resulting \emph{drift matrix} captures behavioral sensitivity across prompts, highlighting instability rooted in tokenization and decoding. Unlike accuracy metrics, PBSS foregrounds rhetorical consistency: do small token-level changes destabilize the output, even when meaning stays the same?

Formal definitions and implementation details are in Appendix~\ref{appendix:pbss}.

\subsection{Projecting Drift: S-BERT as a Behavioral Lens}
PBSS quantifies output consistency by projecting model responses into a semantic embedding space using S-BERT, which captures sentence-level meaning beyond surface variation. This lets us assess whether reworded prompts—identical in intent—elicit stylistically coherent responses.
Viewed through this lens, prompt rephrasing acts as a lightweight perturbation on the token sequence. If such surface-level token shifts produce large embedding divergence, we interpret this as behavioral instability. PBSS thus reframes semantic embeddings as drift detectors—capturing tone, structure, or rhetorical differences without requiring ground-truth labels.
While current benchmarks focus on correctness, PBSS instead probes the internal consistency of model behavior under minor token-level changes. The use of cosine distance provides a simple, interpretable metric for behavioral variability across prompts, temperatures, and models. Ultimately, PBSS asks a simple yet revealing question: \textit{Does the model behave the same when we say the same thing differently?}

\begin{figure*}[!t]
    \centering
    \includegraphics[width=0.94\linewidth]{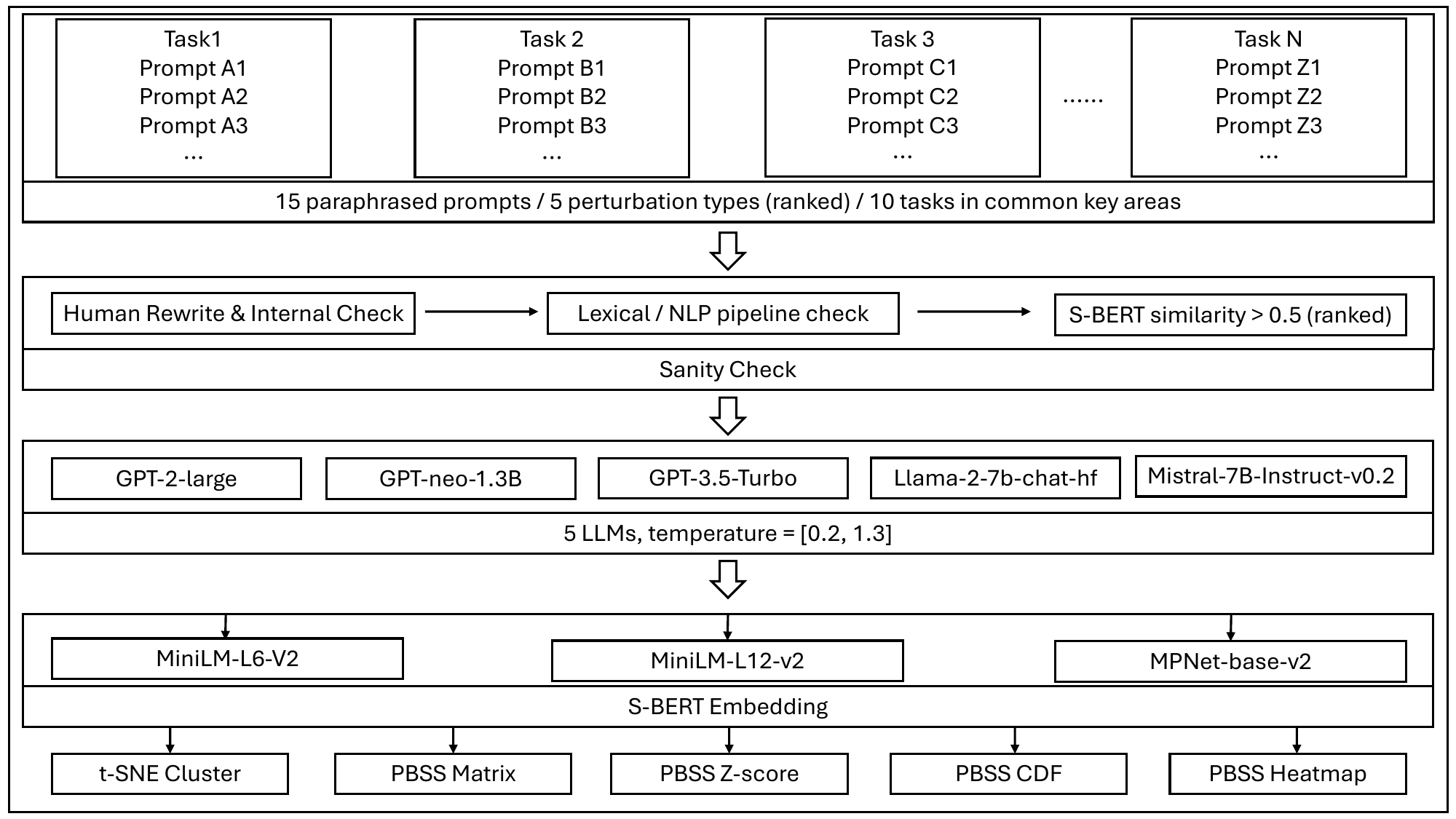}
    \caption{
    Overview of our experimental framework. Structured prompt sets are constructed across ten tasks, undergo multi-layered semantic and lexical validation, and are fed into five LLMs under two decoding temperatures. Outputs are embedded via three S-BERT encoders and analyzed using t-SNE, PBSS matrix, z-score drift detection, and cumulative distribution. This design enables controlled behavioral drift analysis under paraphrastic stress.
    }
    \label{fig:overview_diagram}
\end{figure*}

\section{Experimental Setup}

We evaluate model sensitivity to prompt tokenization by constructing controlled prompt variants across ten task settings. Each task includes a canonical prompt and a set of paraphrased rewordings with minimal semantic drift but varied surface form. These variants induce different tokenization paths while preserving intent, enabling structured analysis of token-level behavioral instability.

Figure~\ref{fig:overview_diagram} presents an overview of the pipeline, and full dataset, model, and decoding configurations are detailed in Appendix~\ref{appendix:setup}.

\subsection{Overview of Evaluation Pipeline}

Figure~\ref{fig:overview_diagram} presents our evaluation workflow. We apply structured prompt perturbations across 10 realistic tasks and evaluate behavioral drift across 5 language models under two decoding temperatures. Outputs are embedded via S-BERT variants and analyzed using PBSS drift scores, cumulative distributions, and visualization metrics (e.g., t-SNE, z-score heatmaps). 

Our goal is to assess whether small lexical or syntactic rewordings—i.e., token-level surface changes—trigger meaningful output drift, even when semantic intent remains constant.

\subsection{Validating Prompt Semantics}

To ensure surface variation without altering intent, each prompt set undergoes three sanity layers: manual review, rule-based syntax filtering, and semantic thresholding using \texttt{all-mpnet-base-v2}. Figure~\ref{fig:S-BERT-similarity-distribution} confirms tight semantic clustering across all task variants.

Figure~\ref{fig:syntax-S-BERT-map} shows example embeddings from two tasks, plotted by S-BERT similarity (x-axis) and shallow syntax distance (y-axis). Prompts shift stylistically, but meaning is preserved—enabling controlled analysis of tokenization-induced drift.

\subsection{Tokenization-Specific Variation Dimensions}

Table~\ref{tab:perturbation-dimensions} defines our perturbation dimensions, each designed to modulate token-level realization while preserving core meaning. This includes syntactic restructuring, tone/style shifts, and malformed input stress tests.

\subsection{Task Coverage}

Table~\ref{tab:prompt-cases} summarizes the 10 task domains. Prompt sets simulate real-world interaction styles—medical guidance, financial justifications, policy explanations—providing a broad substrate for probing token-sensitive behavioral divergence.

\begin{figure}[!t]
    \centering
    \begin{minipage}[t]{1.0\linewidth}
        \centering
        \includegraphics[width=\linewidth]{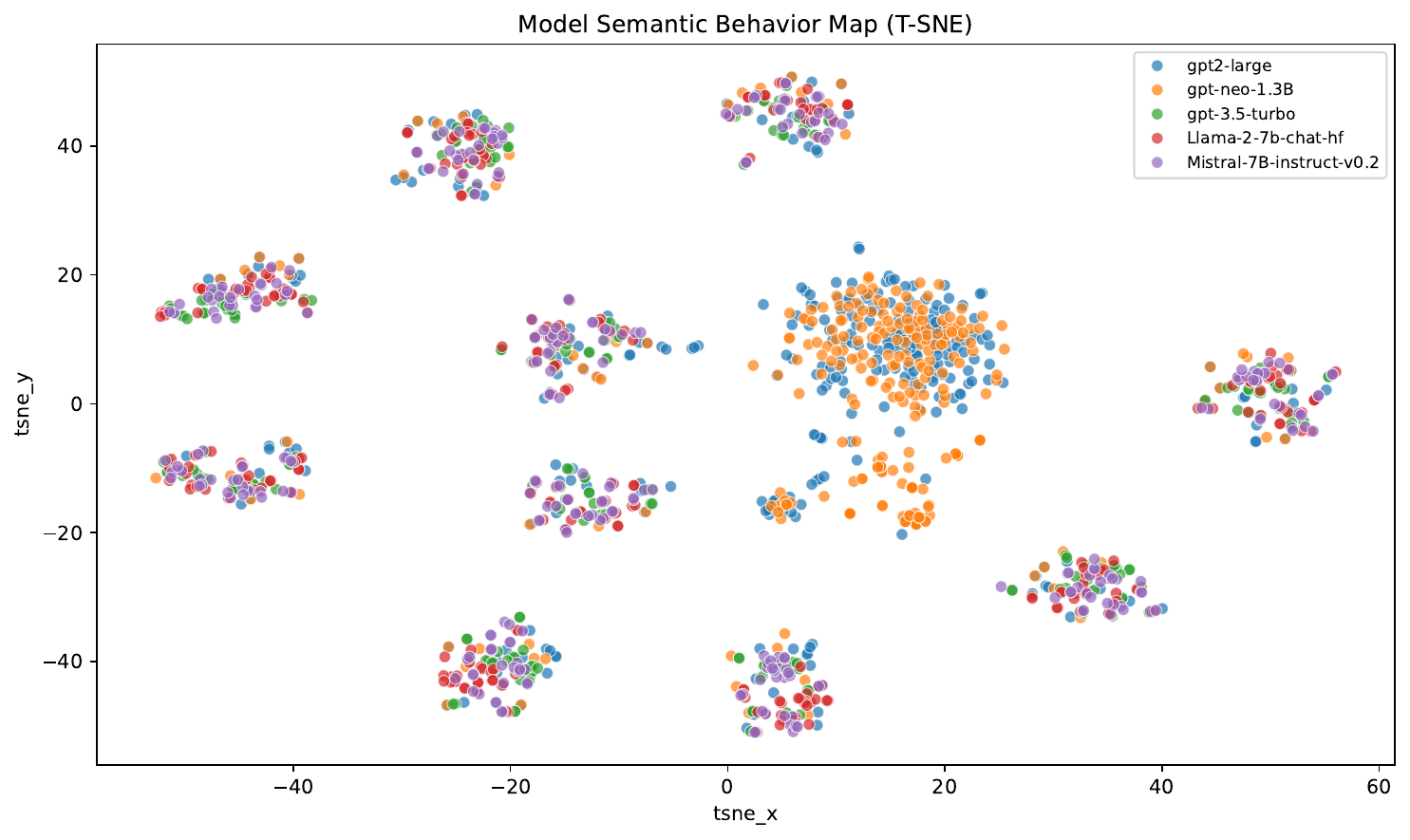}
         \caption{Model-space t-SNE (10 tasks, 5 models) - 1}
        \label{fig:tsne-10p-1a}
    \end{minipage}
    \hfill
    \begin{minipage}[t]{1.0\linewidth}
        \centering
        \includegraphics[width=\linewidth]{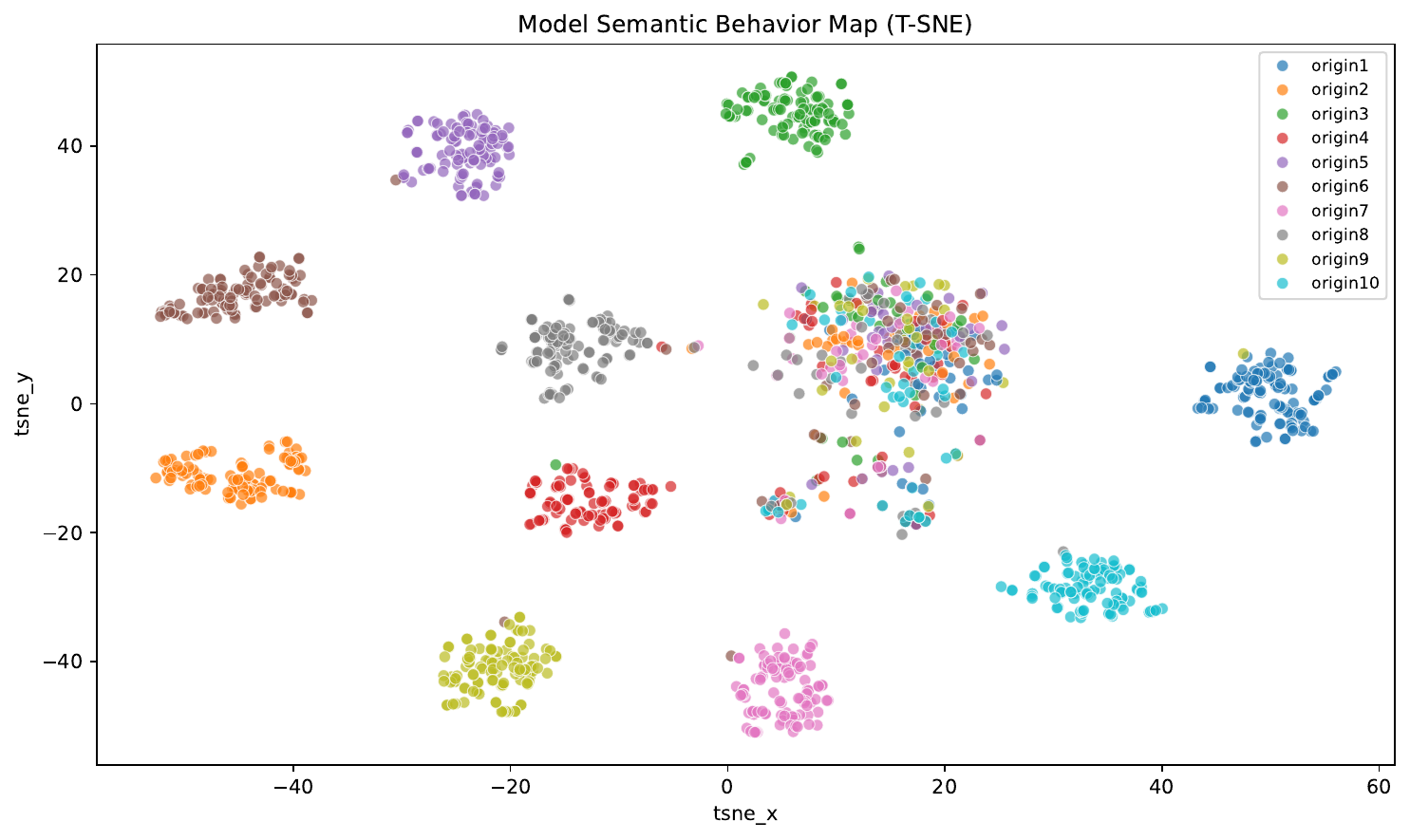}
        \caption{Origin-space t-SNE (10 tasks, 5 models) - 1}
        \label{fig:tsne-10p-2a}
    \end{minipage}
\end{figure}
\section{Patterns and Empirical Findings}\label{sec:experiments}
\subsection{Emergent Task Structure in Model Outputs via Embedding-Space Clustering}

\begin{figure}[!t]
    \centering
    \begin{minipage}[t]{1.0\linewidth}
        \centering
        \includegraphics[width=\linewidth]{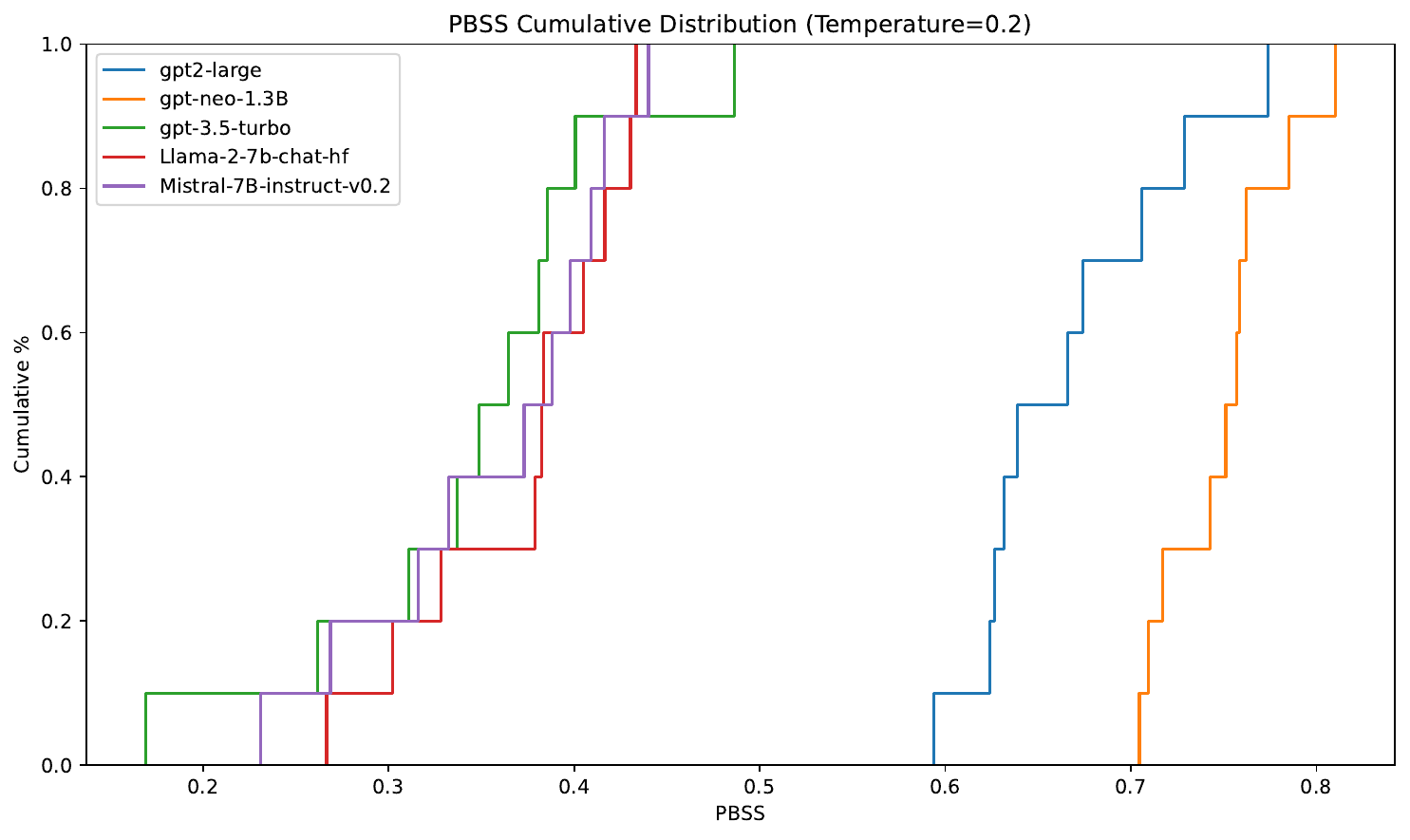}
         \caption{PBSS CDF for MiniLM-L6 under $T=0.2$.}
        \label{fig:pbss-temp02-5}
    \end{minipage}
    \hfill
    \begin{minipage}[t]{1.0\linewidth}
        \centering
        \includegraphics[width=\linewidth]{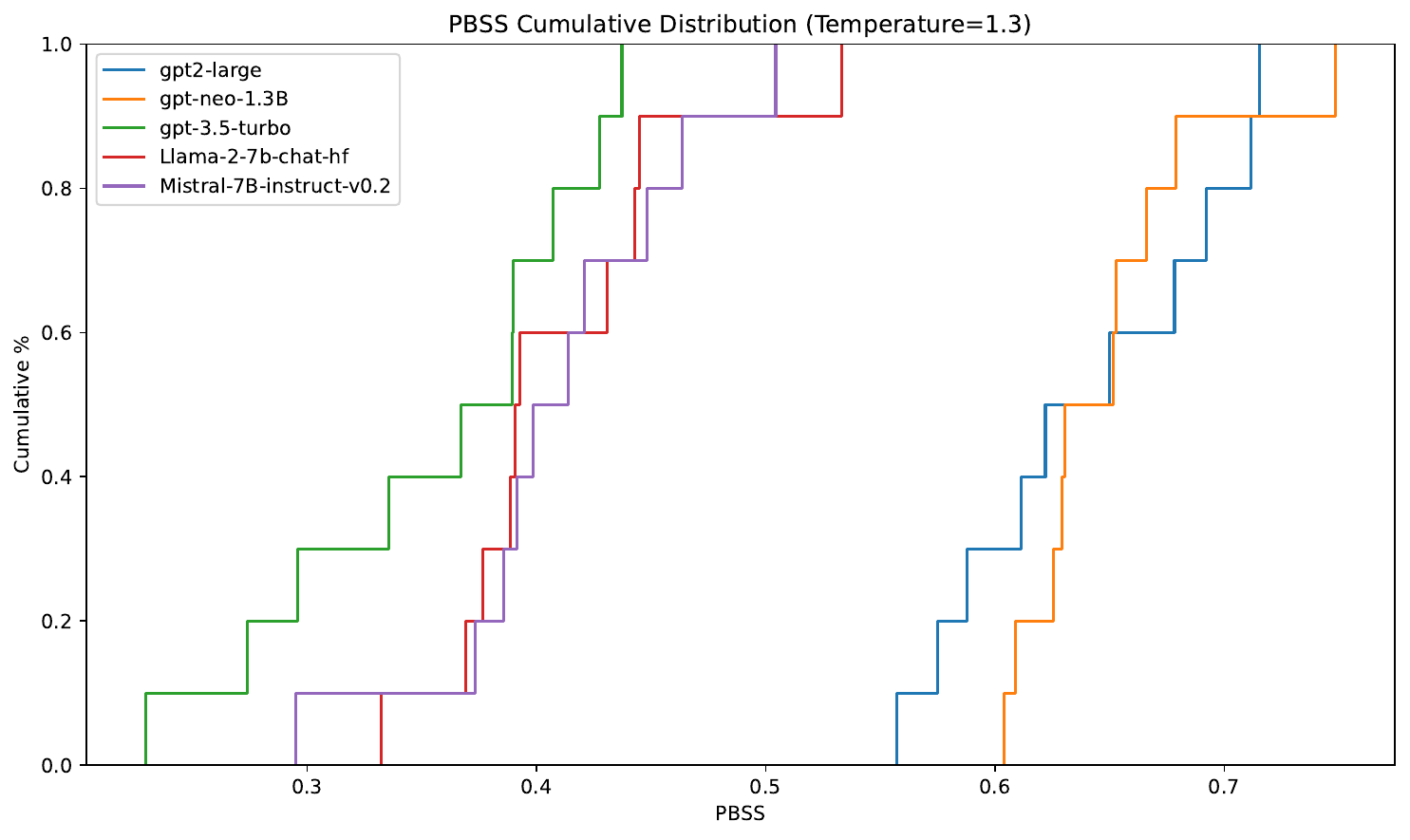}
         \caption{PBSS CDF for MiniLM-L6 under $T=1.3$.}
        \label{fig:pbss-temp13-5}
    \end{minipage}
\end{figure}

To probe latent structure, we embed all model outputs with SBERT (\texttt{all-mpnet-base-v2}) and reduce dimensionality via t-SNE. Across both 3-task and 10-task setups (Fig.~\ref{fig:tsne-10p-1a}–\ref{fig:tsne-10p-2b}), outputs form distinct clusters by task—even without access to origin labels. This suggests that task framing leaves a detectable semantic imprint, offering unsupervised validation of our manual task definitions.

Beyond origin-based clustering, we observe a recurring central cluster across tasks and models—reflecting low-variance, template-like responses. This suggests that some paraphrased prompts induce \textit{semantic compression}, where models default to generic instruction-following patterns despite surface variation. Crucially, this behavior persists across decoding temperatures ($T = 0.2$ and $T = 1.3$), pointing to intrinsic response dynamics rather than sampling noise.

\subsection{Justification for SBERT as an Encoder}

We use SBERT not for absolute fidelity, but for its high sensitivity to rhetorical and semantic shifts—ideal for capturing prompt-induced drift. PBSS relies on relative output structure, and SBERT consistently reveals model-specific patterns across variants, models, and decoding settings.

\subsection{Task Validity via Semantic Embedding Consistency}

Although task categories were manually defined, SBERT+t-SNE projections recover them without supervision, validating the internal coherence of our prompt sets. This indicates that task framing shapes not only output content but also rhetorical structure, producing stable and measurable behavioral signatures.
\begin{figure*}[t]
    \centering
    \begin{minipage}[t]{0.32\linewidth}
        \centering
        \includegraphics[width=\linewidth]{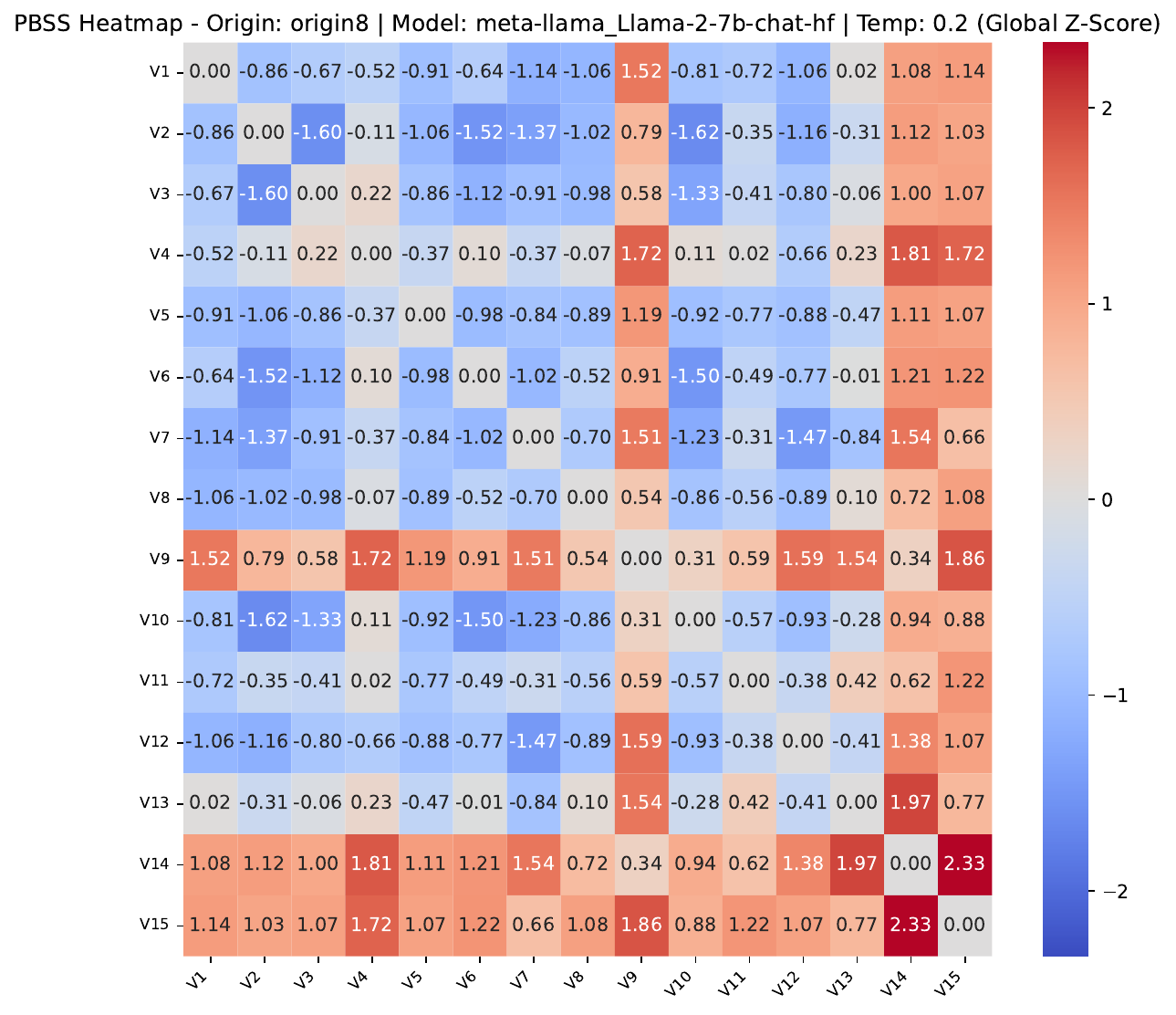}
        \caption{\textbf{MiniLM-L6:} Shows divergent structure and weaker clustering.}
    \end{minipage}
    \hfill
    \begin{minipage}[t]{0.32\linewidth}
        \centering
        \includegraphics[width=\linewidth]{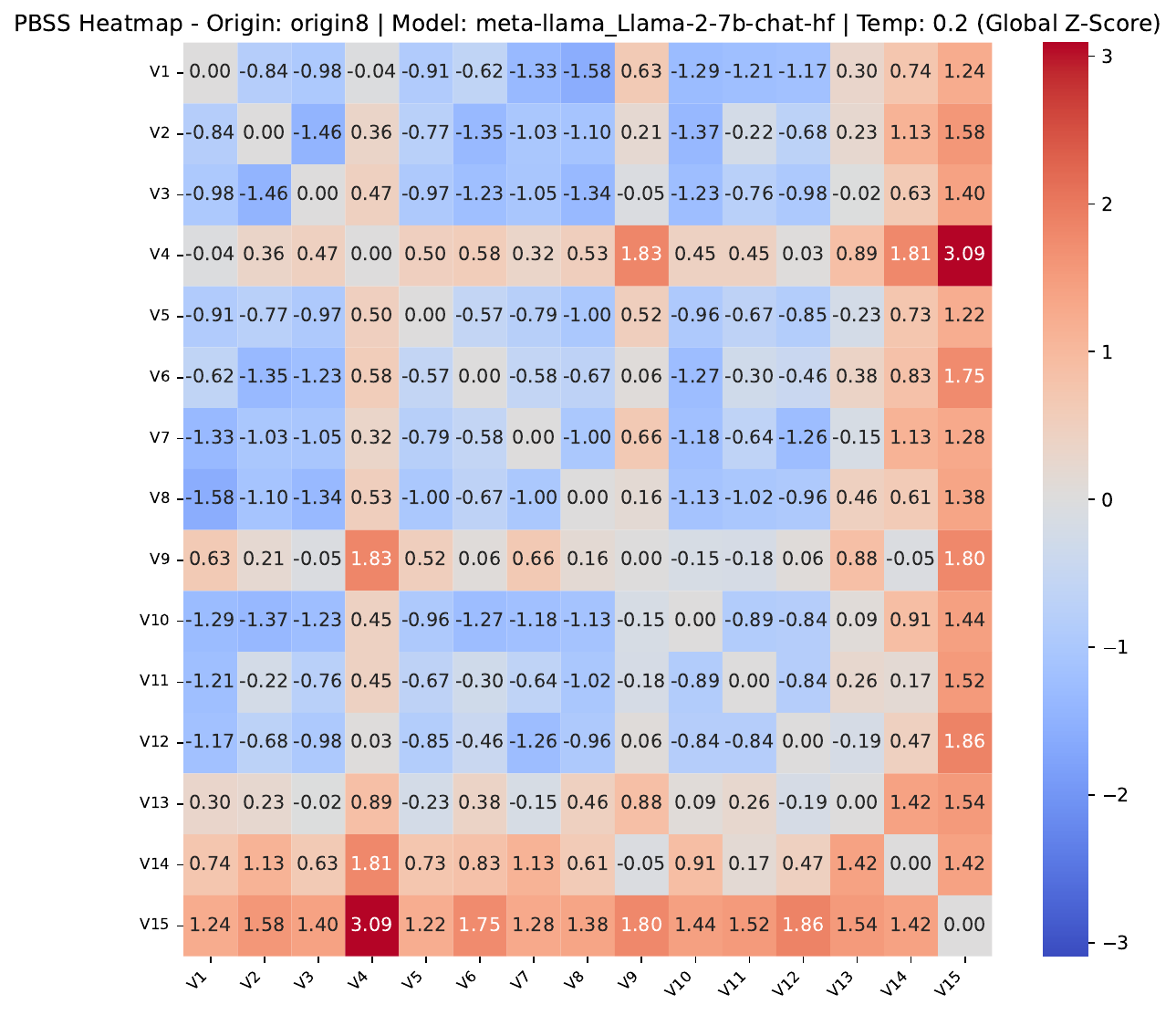}
        \caption{\textbf{MiniLM-L12:} Displays tighter heatmap clusters.}
    \end{minipage}
    \hfill
    \begin{minipage}[t]{0.32\linewidth}
        \centering
        \includegraphics[width=\linewidth]{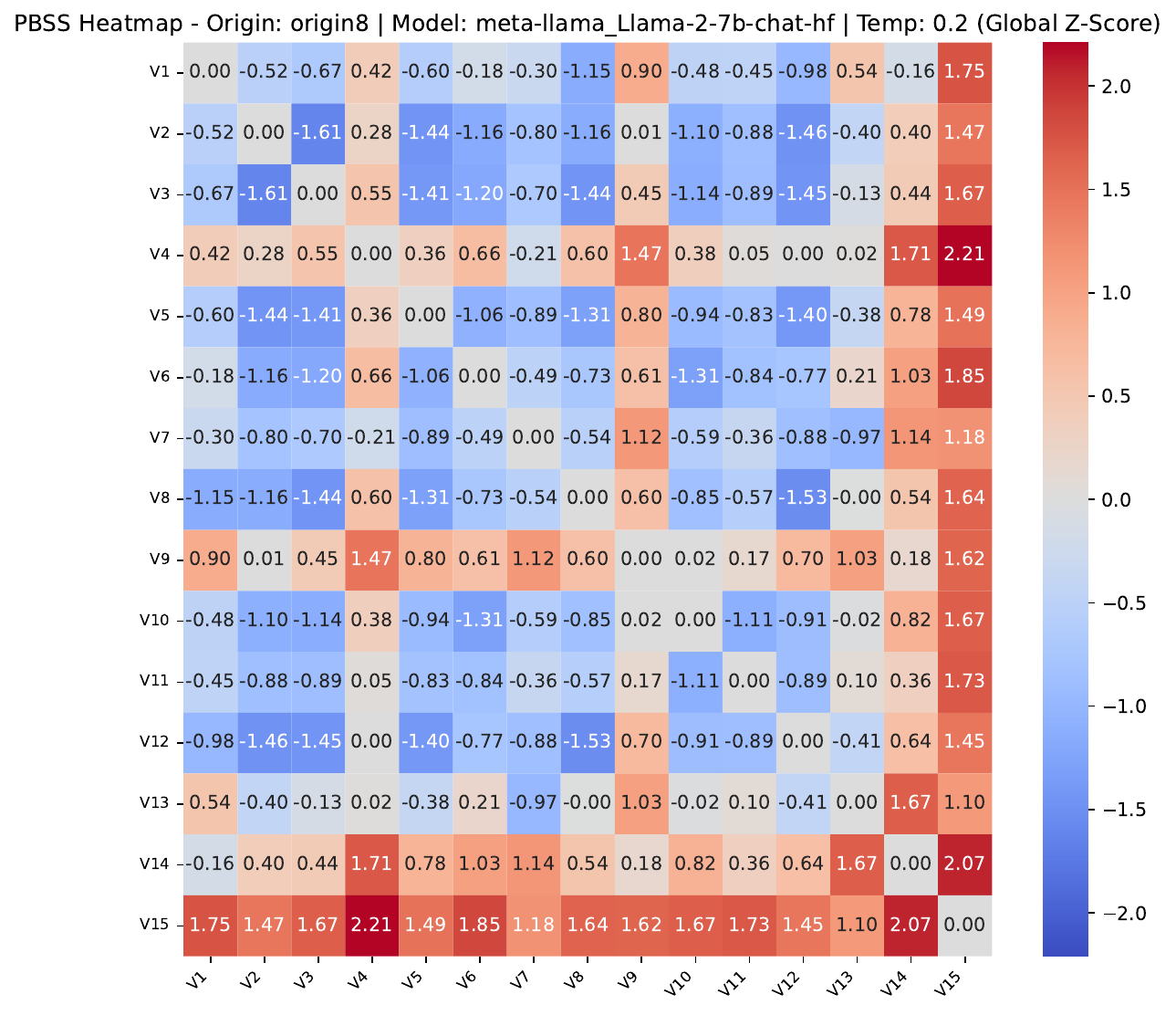}
        \caption{\textbf{MPNet:} Consistent structure across paraphrases.}
    \end{minipage}
\end{figure*}

\subsection{Stable Drift Patterns Across Embedding Architectures}
To test whether PBSS findings depend on the sentence encoder, we compute CDFs of drift scores using three SBERT variants: \texttt{MiniLM-L6}, \texttt{MiniLM-L12}, and \texttt{all-mpnet-base-v2} (Figure~\ref{9-1}-~\ref{9-9}). Across encoders, we observe consistent model rankings and curve shapes: GPT-3.5, LLaMA-2, and Mistral show stable, low-variance behavior; GPT-2 and GPT-Neo display broader drift.

\subsection{Behavioral Drift as a Phase Boundary}
CDF plots reveal a sharp divide: instruction-tuned models (e.g., GPT-3.5, LLaMA-2) cluster tightly under prompt variation, while older models (GPT-2, GPT-Neo) show broad, unstable dispersion. This is not a gradual trend but a behavioral phase shift—newer models adapt phrasing without semantic drift, older ones collapse into repetition or divergence. PBSS thus exposes a clear boundary between alignment-aware and legacy architectures.

\subsection{PBSS Under Variation: Robustness Across Models, Temperatures, and Tasks}

As task coverage expands from 3 to 10 domains, PBSS scores remain stable and consistent, with CDFs smoothing into clear model-specific patterns. Pooling outputs across temperatures ($T = 0.2$, $1.3$) in Figure~\ref{fig:pbss-temp02-5},~\ref{fig:pbss-temp13-5} adds variability but does not disrupt drift structure. These results show that PBSS captures persistent behavioral signals—robust across prompt domains, decoding entropy, and model families.

Across both the 3-task and 10-task settings, we observe strong between-group differences that remain consistent across all SBERT encoder variants, reinforcing the robustness and reliability of the observed drift patterns.

\begin{table*}[htbp]
\centering
\caption{Model Parameter Size Categories}
\label{tab:model-size-categories}
\begin{tabular}{@{}lll@{}}
\toprule
\textbf{Model Name} & \textbf{Parameter Count} & \textbf{Size Category} \\
\midrule
HuggingFaceTB SmolLM-360M & $\sim$0.36B & Small \\
GPT2-Large & $\sim$0.76B & Small \\
EleutherAI GPT-Neo-1.3B & $\sim$1.3B & Small-Medium \\
Microsoft Phi-2 & $\sim$2.7B & Medium \\
Meta-LLaMA Llama-2-7b-chat-hf & 7B & Medium-Large \\
MistralAI Mistral-7B-Instruct-v0.2 & 7B & Medium-Large \\
OpenChat OpenChat-3.5-1210 & $\sim$7B & Medium-Large \\
MythoMax-I2-13B & 13B & Large \\
GPT-3.5-turbo & Undisclosed (API) & Extra-Large \\

\bottomrule
\end{tabular}
\end{table*}

\begin{table*}[ht]
\centering
\caption{Overall PBSS Descriptive Statistics Across All Models}
\label{tab:pbss-summary-total}
\begin{tabular}{@{}lccccc@{}}
\toprule
\textbf{Model} & \textbf{Count} & \textbf{Mean} & \textbf{Std Dev} & \textbf{25\% Quartile} & \textbf{75\% Quartile} \\
\midrule
GPT-3.5-Turbo & 300 & 0.422 & 0.116 & 0.338 & 0.509 \\
Mistral-7B & 300 & 0.427 & 0.103 & 0.351 & 0.504 \\
LLaMA-2-7B & 300 & 0.453 & 0.115 & 0.362 & 0.534 \\
MythoMax-13B & 300 & 0.464 & 0.117 & 0.384 & 0.540 \\
OpenChat-3.5 & 300 & 0.462 & 0.115 & 0.370 & 0.550 \\
Phi-2 & 300 & 0.583 & 0.129 & 0.489 & 0.687 \\
SmolLM-360M & 300 & 0.588 & 0.090 & 0.520 & 0.660 \\
GPT-Neo-1.3B & 300 & 0.650 & 0.103 & 0.566 & 0.736 \\
GPT-2 Large & 300 & 0.635 & 0.094 & 0.563 & 0.705 \\
\bottomrule
\end{tabular}
\end{table*}

\begin{table*}[t]
\centering
\caption{Kruskal–Wallis H-statistics and $p$-values for PBSS across Model Size Groups and SBERT Encoders. Tests are conducted per encoder and aggregated across temperature settings.}
\label{tab:kruskal-summary}
\begin{tabular}{@{}llcccccc@{}}
\toprule
\textbf{Group} & \textbf{Encoder} & \textbf{H (All)} & \textbf{$p$ (All)} & \textbf{H ($T=0.2$)} & \textbf{$p$ ($T=0.2$)} & \textbf{H ($T=1.3$)} & \textbf{$p$ ($T=1.3$)} \\
\midrule
\multirow{3}{*}{\textbf{Small}} 
& MiniLM-L6  & 23.86 & $6.6\times10^{-6}$ & 25.15 & $3.5\times10^{-6}$ & 27.22 & $1.2\times10^{-6}$ \\
& MiniLM-L12 & 18.18 & $1.1\times10^{-4}$ & 37.14 & $8.6\times10^{-9}$ & 20.67 & $3.2\times10^{-5}$ \\
& MPNet      & 25.76 & $2.6\times10^{-6}$ & 50.48 & $1.1\times10^{-11}$ & 15.36 & $4.6\times10^{-4}$ \\
& Combined   & —     & —                  & 103.01 & $4.3\times10^{-23}$ & 31.07 & $1.8\times10^{-7}$ \\
\midrule
\multirow{3}{*}{\textbf{Mid}} 
& MiniLM-L6  & 72.36 & $1.3\times10^{-15}$ & 8.52  & $3.6\times10^{-2}$ & 107.11 & $4.6\times10^{-23}$ \\
& MiniLM-L12 & 100.23 & $1.4\times10^{-21}$ & 21.95 & $6.7\times10^{-5}$ & 111.54 & $5.1\times10^{-24}$ \\
& MPNet      & 67.93 & $1.2\times10^{-14}$ & 7.67  & $5.3\times10^{-2}$ & 97.44  & $5.5\times10^{-21}$ \\
& Combined   & —     & —                   & 34.97 & $1.2\times10^{-7}$ & 306.73 & $3.5\times10^{-66}$ \\
\midrule
\multirow{3}{*}{\textbf{Large}} 
& MiniLM-L6  & 7.79  & $5.3\times10^{-3}$  & 2.78  & $9.5\times10^{-2}$ & 5.24  & $2.2\times10^{-2}$ \\
& MiniLM-L12 & 3.80  & $5.1\times10^{-2}$  & 1.36  & $2.4\times10^{-1}$ & 2.76  & $9.7\times10^{-2}$ \\
& MPNet      & 7.34  & $6.7\times10^{-3}$  & 2.56  & $1.1\times10^{-1}$ & 4.78  & $2.9\times10^{-2}$ \\
& Combined   & —     & —                   & 6.62  & $1.0\times10^{-2}$ & 12.14 & $4.9\times10^{-4}$ \\
\midrule
\multirow{3}{*}{\textbf{All Models}} 
& MiniLM-L6  & 428.89 & $1.2\times10^{-87}$ & 260.94 & $8.2\times10^{-52}$ & 254.38 & $2.0\times10^{-50}$ \\
& MiniLM-L12 & 385.94 & $1.9\times10^{-78}$ & 288.33 & $1.3\times10^{-57}$ & 203.61 & $1.1\times10^{-39}$ \\
& MPNet      & 288.76 & $1.0\times10^{-57}$ & 251.67 & $7.6\times10^{-50}$ & 158.43 & $3.4\times10^{-30}$ \\
& Combined   & —      & —                   & 791.58 & $1.3\times10^{-165}$ & 553.12 & $2.8\times10^{-114}$ \\
\bottomrule
\end{tabular}
\end{table*}

\subsection{Descriptive Statistics.}

To further assess robustness, we expanded the model pool and systematically categorized them by parameter scale (Table~\ref{tab:model-size-categories}), capturing a spectrum from legacy architectures to contemporary instruction-tuned systems. This grouping reflects meaningful architectural and alignment distinctions observed across prior work.

Our extended evaluation spans $10$ diverse tasks, each with $5$ prompt sets and $15$ semantically equivalent paraphrases per set, yielding a total of $4500$ prompt cases per model. While this may appear large, our goal is not to exhaustively probe every response edge case, but to sample sufficient variation to detect emergent patterns. Since PBSS is designed to reveal structural output drift under meaning-preserving rewordings, high coverage improves our ability to identify recurring behavioral signals without requiring massive-scale sweep.

Importantly, our testbed is designed around a controlled explosion of linguistic variation—not to induce overfitting, but to stress-test whether the same semantic input leads to consistent rhetorical behavior across prompt forms. This diversity is crucial to expose latent drift tendencies, especially when comparing encoder outputs across model tiers.

Comprehensive CDF plots and heatmap diagnostics for the full evaluation set are provided in Appendix~\ref{appendix:C}.

We organize models into three empirically motivated groups:

\begin{itemize}
    \item \textbf{Legacy Small Models} (e.g., GPT-2, GPT-Neo-1.3B, SmolLM) — pre-alignment or lightly tuned models with sub-2B parameter counts, showing high variance and poor consistency under paraphrastic perturbation.
    \item \textbf{Mid-Sized, Lightly-Aligned Models} (e.g., LLaMA-2-7B, Phi-2, Mistral-7B, OpenChat) — recent open-source models with alignment objectives but lower capacity or fine-tuning depth, exhibiting intermediate drift behaviors.
    \item \textbf{High-Capacity Instruction-Tuned Models} (e.g., GPT-3.5-Turbo, MythoMax-13B) — commercially-aligned systems with strong consistency across rewordings, indicating drift resistance under prompt variance.
\end{itemize}

This taxonomy reflects both alignment history and capacity tier, and it recapitulates clear behavioral boundaries observed in PBSS distributions. Kruskal–Wallis tests confirm statistically significant divergence between these groups. This structured grouping allows us to compare drift dynamics not just at the model level, but across theoretically relevant behavioral regimes.

Comprehensive CDF visualizations and heatmap diagnostics for all models and encoders are provided in Appendix~\ref{appendix:C}.

\subsection{Significance Testing.}
We apply the Kruskal–Wallis test to PBSS scores across three SBERT encoders and a tiered set of models grouped by size and alignment stage. While a majority of comparisons yield highly significant differences ($p \ll 0.01$), especially between legacy and instruction-tuned systems, we note that significance becomes marginal or disappears in comparisons between models of similar alignment and scale an expected outcome if behavioral convergence is taking place. To assess such patterns structurally, we categorize our model pool into three tiers: (1) \textit{Legacy Small} models such as GPT-2 and GPT-Neo, which exhibit high drift and weak rhetorical consistency under paraphrastic variation; (2) \textit{Mid-Sized Lightly-Aligned} systems, including LLaMA-2 and Phi-2, which demonstrate reduced but still uneven drift patterns; and (3) \textit{High-Capacity Instruction-Tuned} models like GPT-3.5, which show low PBSS variance and limited inter-prompt instability. Importantly, this study is not designed to exhaustively benchmark the space of possible prompts or sampling conditions. Rather, it serves as a structured sanity check: if a well-curated prompt grid can surface interpretable behavioral tiers using only moderate data volume, this implies the presence of latent, internally structured model sensitivity. We report results from a recent evaluation slice spanning 10 tasks, 5 prompt sets per task, 15 semantic variants per set, and 2 decoding temperatures—totaling over 13,000 prompt-response cases per model. Crucially, the PBSS signal remains robust across encoders: despite small differences in absolute score magnitudes, the relative ranking of models remains consistent. This encoder-agnostic stability—what we refer to as \textit{semantic resonance}—suggests that PBSS captures a real behavioral property of models rather than noise induced by embeddings or prompt structure. Together, these findings confirm that semantic drift is measurable, structured, and model-specific—and that PBSS, even under constrained evaluation, provides a reliable lens into this latent axis of LLM behavior.

\subsection{Statistical Summary of Token-Level Drift}

Using multiple methods—Kruskal–Wallis tests, cumulative distributions, and t-SNE projections, we find consistent evidence of behavioral drift triggered by surface-level prompt changes. These effects persist across SBERT variants, decoding temperatures, and prompt domains.

Although each PBSS score reflects a specific pairwise prompt comparison, the construction of prompts was deliberately diversified across tasks, paraphrase templates, and lexical realizations. The 750 unique instructions per model yield 300 distinct prompt pairs without reusing prompt stems—minimizing lexical memorization effects and surface overfitting.

This consistency—across encoder architectures and temperature regimes—suggests that, PBSS captures model-internal regularities rather than spurious correlations induced by prompt design.

Crucially, this drift is not noise: it reflects structured, model-specific sensitivity to the token form of instructions. Even when meaning stays fixed, different surface realizations, often at the token level—can lead to distinct model behaviors. PBSS makes this variation visible, providing a practical diagnostic lens for evaluating how consistently a model responds to semantically equivalent but token-divergent prompts. Were PBSS merely reflecting prompt design idiosyncrasies, we would expect encoder disagreement and temperature instability, neither of which are observed.

\section{QoS-Oriented Evaluation via PBSS}

Having established that prompt variance reflects structured behavioral dynamics rather than incidental noise, we now introduce a practical framework for systematically evaluating model responses to semantically equivalent prompt rewordings. Our approach, Prompt-Based Semantic Shift (PBSS), can be interpreted as a lightweight diagnostic for \textit{Quality-of-Service} (QoS) consistency in large language models (LLMs).

\subsection{Motivation: From Noise to Structured Variance}

Previous analyses—including statistical tests and t-SNE projections, show that model responses to paraphrased prompts do not scatter randomly but instead form structured clusters, indicating systematic behavioral drift. PBSS reframes these observations through a Quality-of-Service (QoS) lens, shifting the focus from correctness to behavioral reliability: how consistently does a model respond when the underlying semantic intent remains unchanged?

\subsection{PBSS as a QoS Metric}

Traditional evaluation frameworks often prioritize task accuracy, but consistent behavior under prompt rewordings is equally critical for the safe and reliable deployment of language models. PBSS metrics and heatmaps offer a practical and interpretable interface for assessing this consistency, addressing a gap in current LLM diagnostics. For instance, when a model is asked to describe the same medical ambiguity using five slightly different prompts, why do some responses converge into fixed phrasing while others vary widely in tone and structure?

PBSS makes this variation visible, traceable, and, importantly, diagnosable. This has significant implications for domains like clinical NLP, where LLMs are increasingly tasked with generating summaries, treatment recommendations, or patient-facing explanations. In these contexts, narrative structure is tightly linked to interpretation: subtle changes in style or emphasis can shape downstream clinical decisions. By quantifying behavioral consistency under controlled paraphrastic shifts, PBSS provides a critical diagnostic layer for evaluating the robustness of medical language models.

Notably, we observe that different SBERT encoders produce distinct PBSS heatmap structures under global z-score normalization. This highlights PBSS’s sensitivity not only to output variability but also to the semantic granularity introduced by the choice of embedding model. Such variation is expected, as SBERT encoders differ in how they model sentence-level similarity, shaped by differences in training objectives (e.g., natural language inference vs. semantic textual similarity), architectural depth, and embedding scale sensitivity.

\section{Discussion}
\subsection{Implications and Boundaries}

This study does not aim to simulate identity, personality, or anthropomorphic consistency. All observations are derived from neutral, task-oriented prompts within the standard instruction-following framework, with no use of adversarial or fine-tuned inputs.

Crucially, the drift patterns we observe are not isolated anomalies, but consistent and statistically verifiable trends. Their emergence from semantically stable, non-adversarial prompts highlights a latent behavioral dimension in language models—one that is measurable, reproducible, and increasingly important for developers to understand.

\subsection{Prompt Drift and Jail-Breaking Behavior}

Jailbreaking can be reframed as an extreme case of prompt-induced drift. Instead of seeing jailbreaks solely as security failures, we interpret them as responses from high-sensitivity regions of the prompt space—zones where minor changes yield disproportionately deviant outputs.

PBSS may help identify these zones early. Though we do not test jailbreak prompts directly, future work could explore whether high PBSS drift correlates with jailbreak susceptibility, allowing proactive model stress-testing from a behavioral perspective.

\subsection{Clinical Relevance and High-Stakes Domains}

Certain downstream domains—particularly clinical and diagnostic settings—are acutely sensitive to stylistic ambiguity and rhetorical inconsistency. Medical prompts tend to be semantically dense, structurally constrained, and highly dependent on context. In such environments, even minor shifts in prompt phrasing can lead to conflicting interpretations, compromising both safety and reliability.

Unlike open-ended creative tasks, where expressive variation is desirable, clinical applications require semantic precision and deterministic responses. In these cases, prompt variance is not just a linguistic anomaly—it represents a potential risk vector.

With a growing interest in incorporating AI systems into the clinical workflow, there is no room for error when entering prompts. Even a slight semantic shift can lead to catastrophic consequences, namely a misinterpretation of clinical data that results in a misdiagnosis and, thus, potentially worse outcomes for patients, especially in settings requiring immediate action, e.g., the emergency room. In such high-stakes environments, precision in communication is critical; the margin for ambiguity is virtually nonexistent. An AI-generated electronic medical record (EMR) with errors or misinterpreted recommendations could delay treatment, misguide interventions, or obscure vital diagnostic signals. Therefore, ensuring consistency and clarity in how prompts are constructed and how models respond is not simply a technical concern—it is a clinical imperative. The integration of AI in emergency medicine, diagnostics, and decision support must be accompanied by a rigorous validation of its semantic stability under variable phrasing to protect patient outcomes and uphold the standard of care.

Here, PBSS offers a low-cost front-end screening tool: it flags reactivity under benign variations before deployment in real-world, high-stakes workflows. As LLMs move closer to medical decision support, prompt variance could emerge as a risk vector, where surface-level phrasing differences yield divergent outputs under stable intent.

\subsection{Pattern, Not Noise}

PBSS does not evaluate what models say, but how they say it, offering a minimal lens on rhetorical consistency under paraphrastic variation. Our results show structured, repeatable behavioral transitions across rewordings. Models display unique rhetorical response patterns, consistent within model families but sensitive to phrasing shifts.

We make no anthropomorphic claims. But we do observe model-specific behavioral landscapes—semantic surfaces where paraphrases induce predictable rhetorical divergence. Understanding this structure is essential for interpretability, deployment, and interface robustness.

\section{Future Directions}

\paragraph{PBSS as a Behavioral Lens}
We frame PBSS not as a benchmark, but as a minimal probe for studying LLM consistency under semantically fixed, stylistically varied prompts. Our results suggest such variation reflects structured behavioral drift—systematic, not incidental.

\paragraph{Token-Level Drift and Latent Inertia}
Drift often emerges from differences in phrasing or tone—suggesting persistent stylistic priors shaped by pretraining. This inertia may reflect deeper prompt-processing biases. Studying it in multi-turn or role-adaptive settings could expose how models encode rhetorical context over time.

\paragraph{Drift as a Jailbreak Signal}
PBSS offers a non-adversarial lens on jailbreak phenomena. High drift regions in prompt space may align with instability thresholds that models exploit or misinterpret. This opens new paths for jailbreak risk diagnostics without relying on exploit engineering.

\paragraph{Safety-Critical Deployment}
In domains like medicine or policy, rhetorical inconsistency is a risk vector. PBSS can serve as a lightweight front-end filter, flagging instability in prompts before models are deployed in high-stakes scenarios.

\paragraph{Toward Interpretability}
Prompt rewording activates distinct pathways through the model’s tokenization and decoding landscape. PBSS can help connect surface behavior to internal dynamics—enabling cognitive and interpretive insights into how LLMs construct meaning.

\paragraph{Open Framework}
We release PBSS as a minimal, extensible diagnostic scaffold—inviting the community to extend it as a foundation for studying token-induced behavioral variance beyond correctness, into consistency and rhetorical alignment.

\bibliography{main}

\newpage
\appendix
\label{appendix:pbss}
\onecolumn
\section{Formal Definition of PBSS Framework}
 Let $\mathcal{P}$ denote a set of paraphrastic prompts for a fixed task. Let $f$ be a language model, such that $f: \mathcal{P} \rightarrow \mathcal{Y}$, mapping inputs to textual outputs. For prompts $p_i, p_j \in \mathcal{P}$, let $y_i = f(p_i)$ and $y_j = f(p_j)$ be the generated responses.

Let $s: \mathcal{Y} \rightarrow \mathbb{R}^d$ denote an embedding function (e.g., S-BERT). We define a drift indicator $D$ as the cosine distance between outputs:

\[
D(p_i, p_j) = 1 - \cos(s(y_i), s(y_j)) = 1 - \frac{s(y_i) \cdot s(y_j)}{\|s(y_i)\| \cdot \|s(y_j)\|}
\]

The drift score quantifies the extent to which a model's outputs vary in tone, structure, or style when presented with prompts that differ in surface form but preserve semantic intent. Within the PBSS framework, we calculate this score $D$ across all pairs of paraphrased prompts, assembling the results into a diagnostic drift matrix that captures the model’s sensitivity profile across variations.

\paragraph{PBSS Cumulative Distribution (CDF).}
Given a set of $n$ paraphrased prompts $\{p_1, ..., p_n\}$, we compute all pairwise PBSS drift scores using a semantic embedding function $s(\cdot)$:

\[
D_{i,j} = 1 - \cos(s(f(p_i)), s(f(p_j))), \quad \forall i \ne j
\]

This results in a set of $\binom{n}{2}$ drift scores across the prompt set. We define the PBSS \textbf{Cumulative Distribution Function (CDF)} as:

\[
F(\delta) = \frac{1}{|S|} \sum_{(i,j) \in S} \mathbb{I}[D_{i,j} \le \delta]
\]

where $S = \{(i,j) \mid i < j\}$ and $\delta \in [0, 2]$ (max cosine distance range). $F(\delta)$ represents the proportion of prompt pairs with behavioral divergence less than or equal to threshold $\delta$.

This distribution captures the model’s overall stability in response to surface-level prompt variation. A steep cumulative distribution function (CDF) that rises sharply near $\delta = 0$ indicates high behavioral consistency, while a flatter or right-shifted curve reflects greater variability or more frequent drift across paraphrased prompts. In the PBSS framework, we quantify this variability by computing the drift score 
$D$ across all prompt pairs, assembling the results into a diagnostic drift matrix that summarizes the model’s sensitivity profile.

\paragraph{Optional Hybrid Formulation.} In settings where semantic equivalence is imperfect or uncertain, we introduce a hybrid score incorporating semantic similarity:

\[
\text{PBSS}_{\text{hybrid}}(p_i, p_j) = \lambda \cdot \text{Sim}_{\text{sem}}(y_i, y_j) + (1 - \lambda) \cdot \text{PBSS}(p_i, p_j)
\]

where $\text{Sim}_{\text{sem}}$ denotes a semantic similarity measure (e.g., S-BERT cosine similarity), and $\lambda \in [0,1]$ controls the trade-off between semantic preservation and stylistic consistency. We note that $\text{Sim}_{\text{sem}}$ is one possible instantiation, and not intrinsic to PBSS.

\paragraph{PBSS Matrix.} Given $n$ prompts $\{p_1, \dots, p_n\}$, we compute the full PBSS matrix $D \in \mathbb{R}^{n \times n}$, where $D_{i,j} = \text{PBSS}(p_i, p_j)$ for $i \neq j$ and $D_{i,i} = 0$.

From this matrix, we derive:

\begin{itemize}
    \item Mean PBSS: $\mu = \frac{1}{n(n-1)} \sum_{i \ne j} D_{i,j}$
    \item Max PBSS: $\max_{i \ne j} D_{i,j}$
    \item CDF curve: empirical distribution of PBSS values over all $(i,j)$ pairs
\end{itemize}

\paragraph{Z-score Drift Indexing.} To detect salient behavioral outliers, we define two z-scoring methods:

\textit{(1) Global Z-score} over all off-diagonal entries:

\[
Z_{i,j}^{\text{global}} = \frac{D_{i,j} - \mu_D}{\sigma_D}, \quad i \ne j
\]

where $\mu_D$ and $\sigma_D$ are the global mean and standard deviation of $D$ over $i \ne j$.

\textit{(2)Row-normalized Z-score}, measuring deviation relative to each prompt’s neighborhood:

\[
Z_{i,j}^{\text{row}} = \frac{D_{i,j} - \mu_i}{\sigma_i}, \quad i \ne j
\]

where $\mu_i$, $\sigma_i$ are the mean and std of row $i$ excluding the diagonal.

These indices allow localized drift patterns to emerge, highlighting prompts whose phrasing induces outsized behavioral responses.

\section{Experiment Setup Details}
\label{appendix:setup}
\subsection{Model Configurations}
We evaluate five large language models: GPT-2 (774M), GPT-Neo (1.3B), LLaMA-2 (7B), Mistral (7B), and GPT-3.5-Turbo (via OpenAI API). Models are queried using identical prompt batches under two decoding temperatures: $T=0.2$ (low-variance) and $T=1.3$ (high-variance). This setup probes behavioral consistency under both stable and stochastic generation regimes.

Open-weight models are accessed via HuggingFace Transformers with greedy or top-k sampling (as applicable). GPT-3.5-Turbo is accessed using the OpenAI Platform API v1.

\subsection{Sentence Embedding Models}

To compute PBSS drift scores, we embed generated outputs using three Sentence-BERT (S-BERT) variants:

\begin{itemize}
    \item \textbf{MiniLM-L6-v2}: compact 6-layer transformer, optimized for fast inference;
    \item \textbf{MiniLM-L12-v2}: deeper version with increased representational granularity;
    \item \textbf{all-mpnet-base-v2}: pre-trained via permuted language modeling; used for filtering and validation.
\end{itemize}

These models vary not only in depth and parameter count, but also in their training objectives—for example, masked language modeling versus permuted language modeling. By applying all three encoders in parallel throughout the evaluation pipeline, we assess whether behavioral drift patterns remain consistent across embedding architectures. This serves as a robustness check, minimizing reliance on any single semantic representation.

All embeddings are cosine-normalized and processed with batch inference via the \texttt{sentence-transformers} Python library.

\subsection{Prompt Logging Schema}

All model outputs are captured through a lightweight, modular logging system built for reproducibility and easy extensibility. To enable systematic behavioral analysis, we additionally formalize two structured data schemas: the \textbf{Prompt Variant Logging Schema} and the \textbf{PBSS Evaluation Summary Schema}. These schemas serve as extensible templates for recording prompt perturbations, model parameters, and aggregated drift scores—designed to facilitate downstream diagnostics, comparative evaluation, and integration with future tools.

\begin{table*}[t]
  \centering
  \small 
  \begin{minipage}[t]{0.47\linewidth}
    \centering
    \caption{Prompt Variant Logging Schema}
    \label{tab:appendix-logging}
    \begin{tabular}{@{}ll@{}}
      \toprule
      \textbf{Field}          & \textbf{Description}              \\
      \midrule
      \texttt{origin}         & Original task prompt identifier   \\
      \texttt{variant\_id}    & Prompt variant ID                \\
      \texttt{model\_name}     & Model name (e.g., GPT-2)          \\
      \texttt{temperature}     & Decoding temperature              \\
      \texttt{prompt}          & Prompt string                     \\
      \texttt{output\_text}    & Model-generated output            \\
      \bottomrule
    \end{tabular}
  \end{minipage}\hfill
  \begin{minipage}[t]{0.47\linewidth}
    \centering
    \caption{PBSS Evaluation Summary Schema}
    \label{tab:pbss-logging}
    \begin{tabular}{@{}ll@{}}
      \toprule
      \textbf{Field} & \textbf{Description}                                    \\
      \midrule
      \texttt{origin}      & Original task prompt identifier                        \\
      \texttt{model}       & Language model used for generation                     \\
      \texttt{temperature} & Decoding temperature during inference                  \\
      \texttt{avg\_pbss}   & Mean drift score across all prompt variants            \\
      \bottomrule
    \end{tabular}
  \end{minipage}
\end{table*}

\begin{table*}[t]
  \centering
  \small
  \begin{minipage}[t]{0.47\linewidth}
    \centering
    \caption{Definition of Prompt Perturbation Dimensions}
    \label{tab:perturbation-dimensions}
    \begin{tabular}{@{}p{0.25\linewidth}p{0.6\linewidth}@{}}
      \toprule
      \textbf{Dimension}           & \textbf{Description}                                  \\
      \midrule
      \textbf{Stylistic Shift}     & Varies tone/register (e.g., formal to conversational).      \\
      \textbf{Syntactic Manipulation} & Reorders or modifies grammatical structure.               \\
      \textbf{Instructional Perturbation} 
                                  & Alters framing verbs or phrasing.                        \\
      \textbf{Contextual Reframing} & Wraps tasks in scenario contexts.                         \\
      \textbf{Broken Prompt}       & Injects malformed or compressed input.                  \\
      \bottomrule
    \end{tabular}
  \end{minipage}\hfill
  \begin{minipage}[t]{0.47\linewidth}
    \centering
    \caption{Prompt Cluster Coverage Across Use Domains}
    \label{tab:prompt-cases}
    \begin{tabular}{@{}p{0.12\linewidth}p{0.18\linewidth}p{0.55\linewidth}@{}}
      \toprule
      \textbf{Case ID} & \textbf{Domain}         & \textbf{Task Description}                          \\
      \midrule
      C1  & Medical Explanation  & Explain motion-induced boundary ambiguity in MRI scans     \\
      C2  & Causal Inference     & Identify confounding factors in paradoxical drug effects    \\
      C3  & Urban Policy         & Design fair and effective weekday traffic restriction policy \\
      C4  & Public Science       & Communicate climate effects in varying tone registers         \\
      C5  & Primary Education    & Explain a basic concept in age-appropriate language           \\
      C6  & Academic Appeals     & Justify a request for grade review                            \\
      C7  & Major Transfer       & Motivate an undergraduate program switch convincingly         \\
      C8  & Financial Forms      & Write compliant yet natural reimbursement justifications      \\
      C9  & Audit Response       & Draft formal responses to regulatory inquiries                \\
      C10 & Request Protocol     & Vary formality and tone in formal administrative requests     \\
      \bottomrule
    \end{tabular}
  \end{minipage}
\end{table*}

\subsection{Codebase and Reproducibility}

This structure enables precise tracking of each prompt variant across different models and temperature settings, facilitating aligned behavioral comparisons. 

\newpage
\vspace{1em}

\begin{figure}[htbp]
\centering
\includegraphics[width=0.7\linewidth]{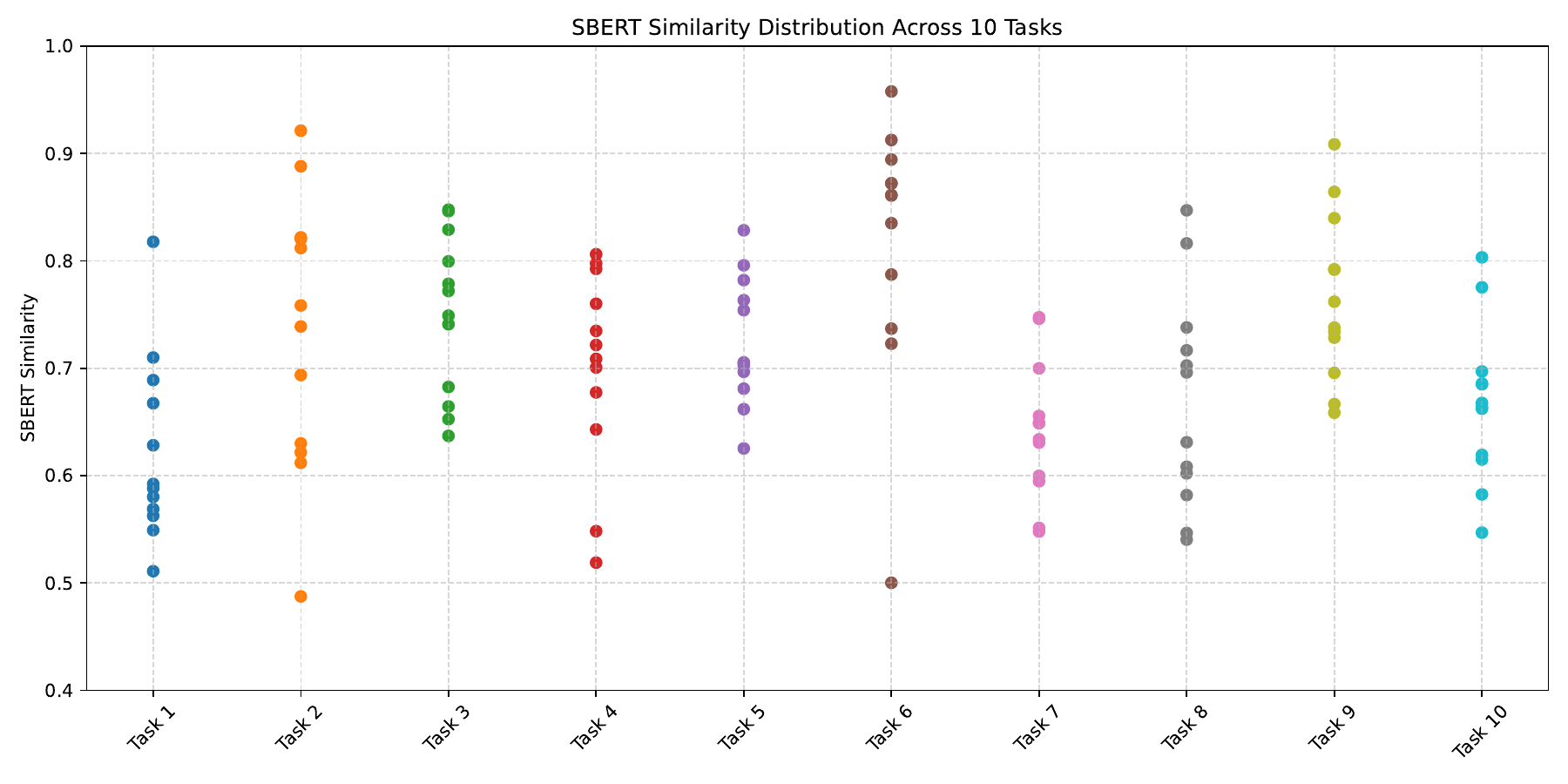}
\caption{S-BERT-based similarity distribution for prompt variants across 10 tasks. Each prompt set per task includes 15 prompt variations semantically aligned with its original instruction.}
\label{fig:S-BERT-similarity-distribution}
\end{figure}

\begin{figure*}[htbp]
  \centering
  \subfigure[C1: Medical Explanation]{
    \includegraphics[width=0.47\textwidth]{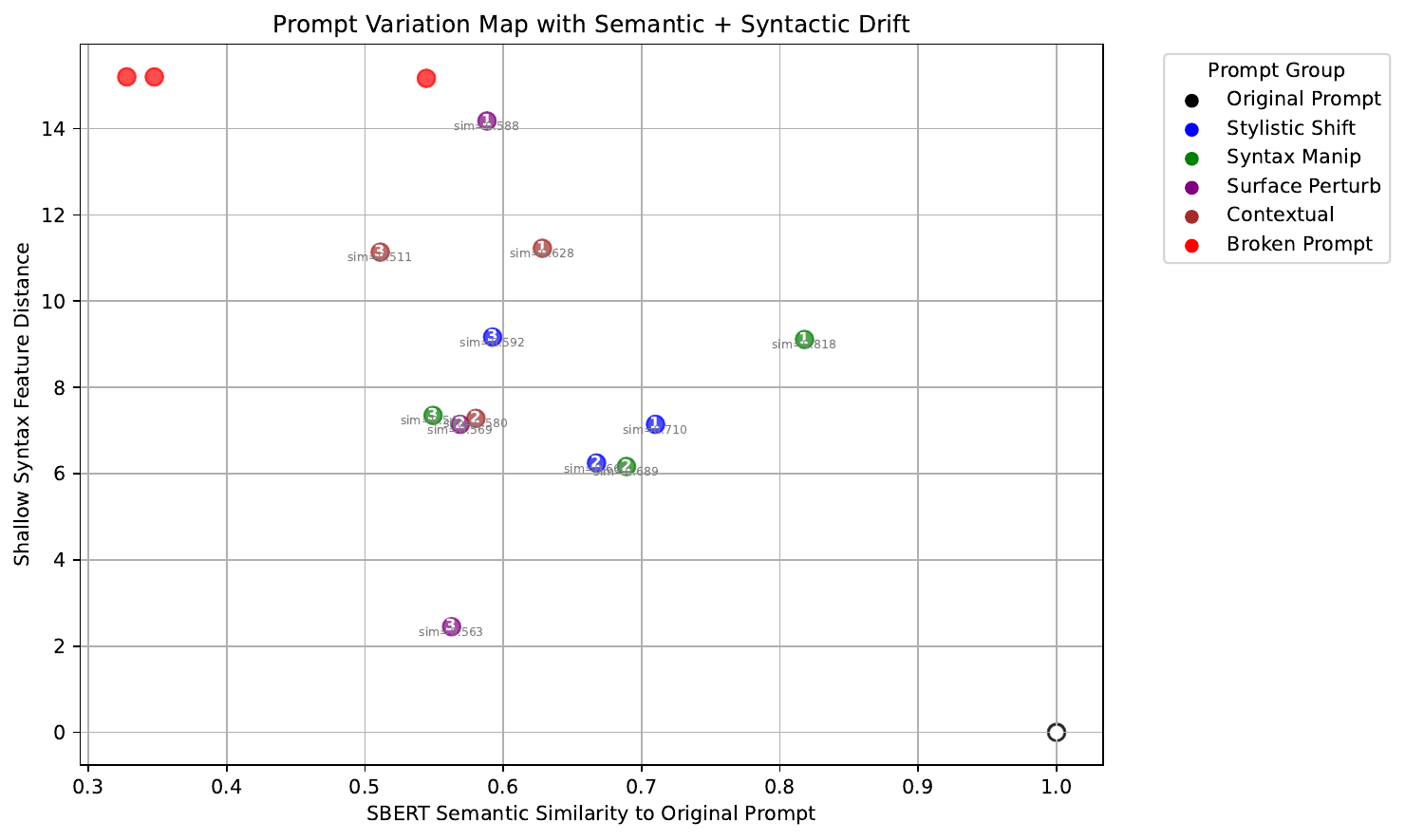}
    \label{fig:map-c1}
  }
  \hfill
  \subfigure[C8: Financial Forms]{
    \includegraphics[width=0.47\textwidth]{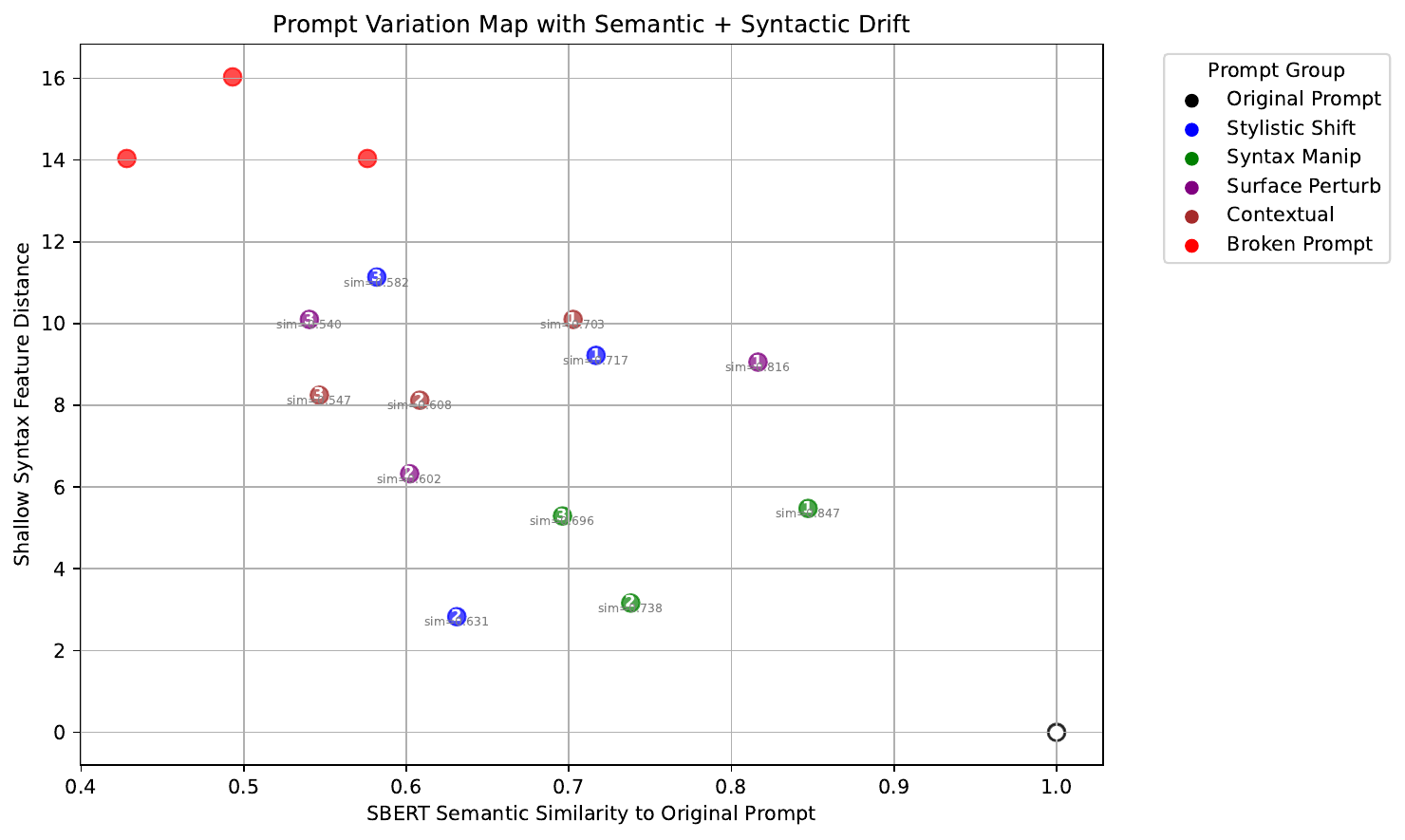}
    \label{fig:map-c8}
  }
\caption{Each point reflects a prompt variant, plotted by S-BERT similarity (x-axis) and shallow syntax distance (y-axis). This sanity check visually confirms that the prompt set maintains semantic consistency while exhibiting token-level surface variation.}
  \label{fig:syntax-S-BERT-map}
\end{figure*}

\section{Extended Visualization of Drift Patterns}
\label{appendix:C}
To support our main analysis, we provide extended visualizations below. In addition to the five core models analyzed in the main text, we introduce five new LLMs to perform cross-validation across architectural families. We visualize drift patterns on both 3-task and 10-task subsets to assess consistency under varying semantic and lexical constraints.

Model outputs are embedded via S-BERT and projected using t-SNE. Plots are organized by model identity (model space) and prompt origin (origin space). The 3-task setup captures finer prompt-level variation, while the 10-task configuration offers broader domain coverage. Visual coherence in origin space reflects output stability across paraphrased prompts; dispersed or fragmented clusters indicate higher behavioral sensitivity.

\begin{figure*}[htbp]
    \centering
    \begin{minipage}[t]{0.48\linewidth}
        \centering
        \includegraphics[width=\linewidth]{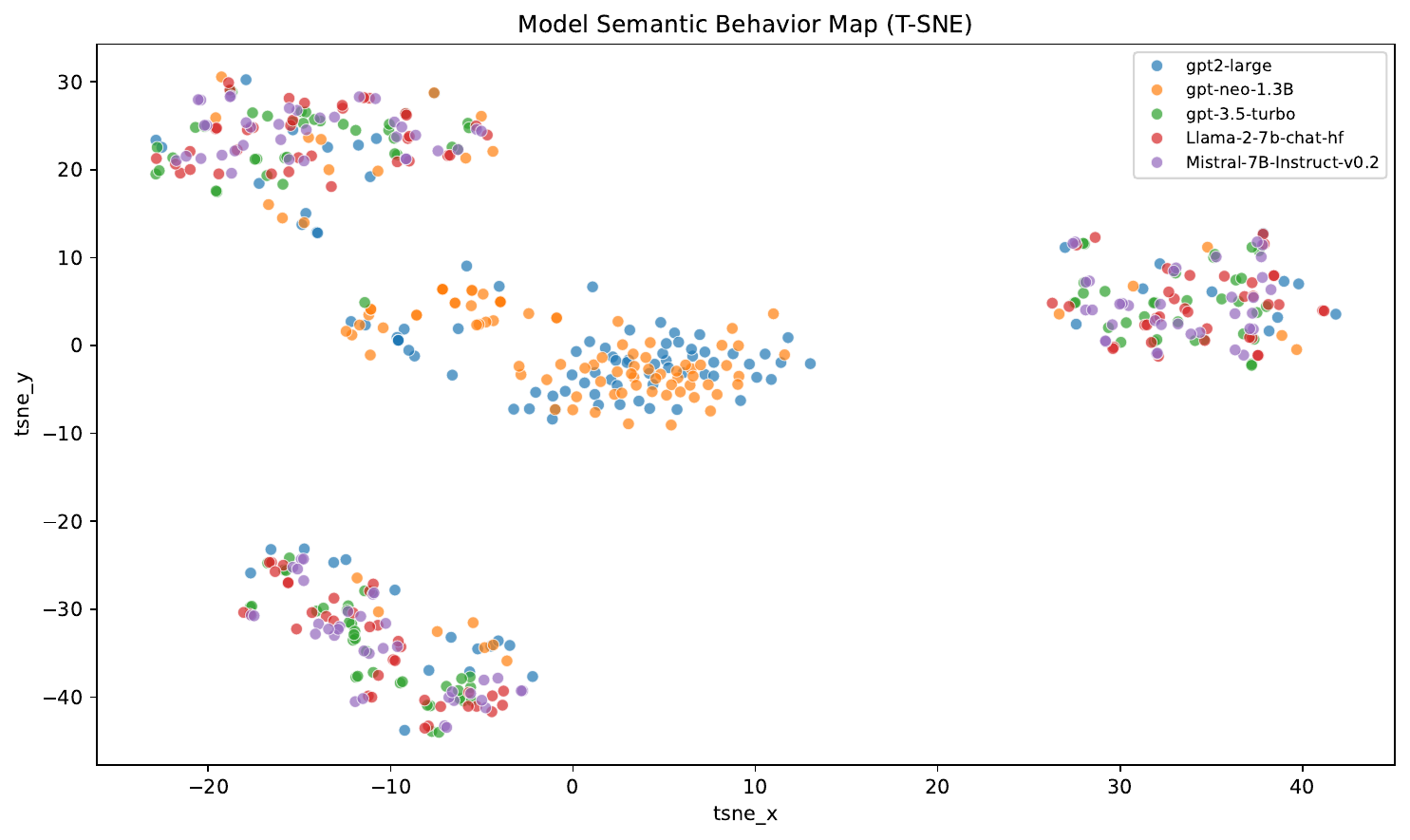}
        \caption{Model-space t-SNE (3 tasks, 5 models)}
        \label{fig:tsne-3p-1}
    \end{minipage}
    \hfill
    \begin{minipage}[t]{0.48\linewidth}
        \centering
        \includegraphics[width=\linewidth]{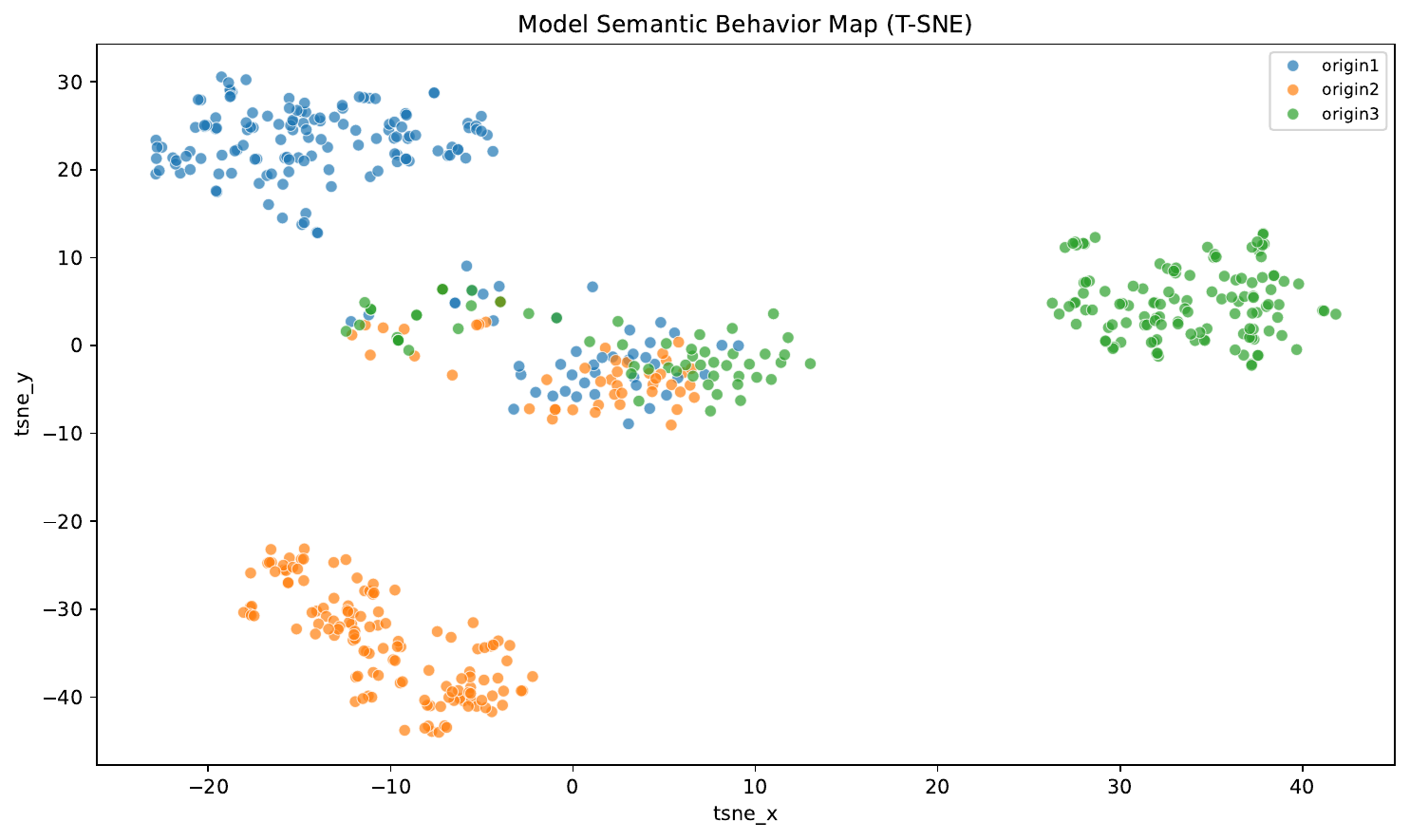}
        \caption{Origin-space t-SNE (3 tasks, 5 models)}
        \label{fig:tsne-3p-2}
    \end{minipage}
     \begin{minipage}[t]{0.48\linewidth}
        \centering
        \includegraphics[width=\linewidth]{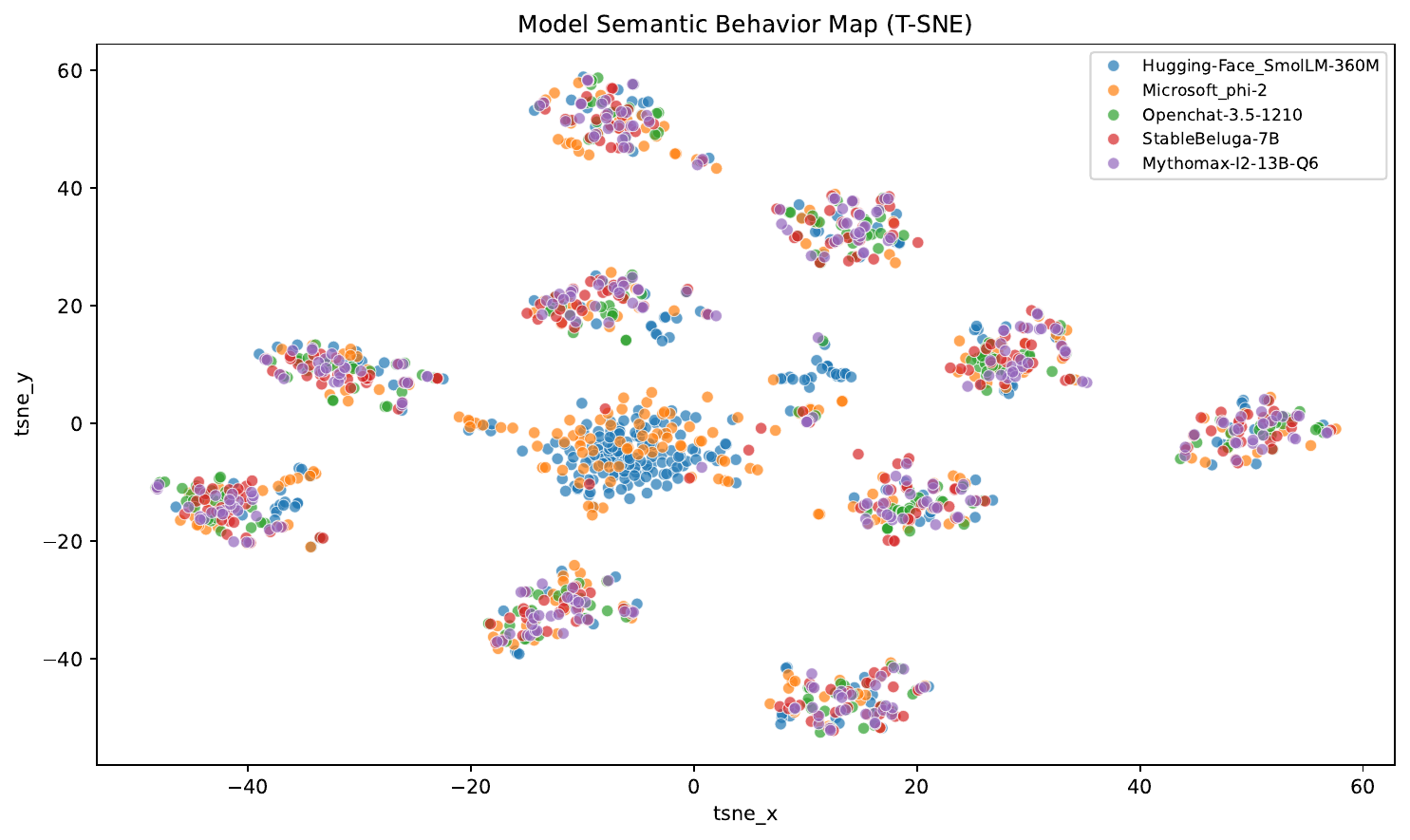}
        \caption{Model-space t-SNE (10 tasks, 5 new models) - 2}
        \label{fig:tsne-10p-1b}
    \end{minipage}
    \hfill
    \begin{minipage}[t]{0.48\linewidth}
        \centering
        \includegraphics[width=\linewidth]{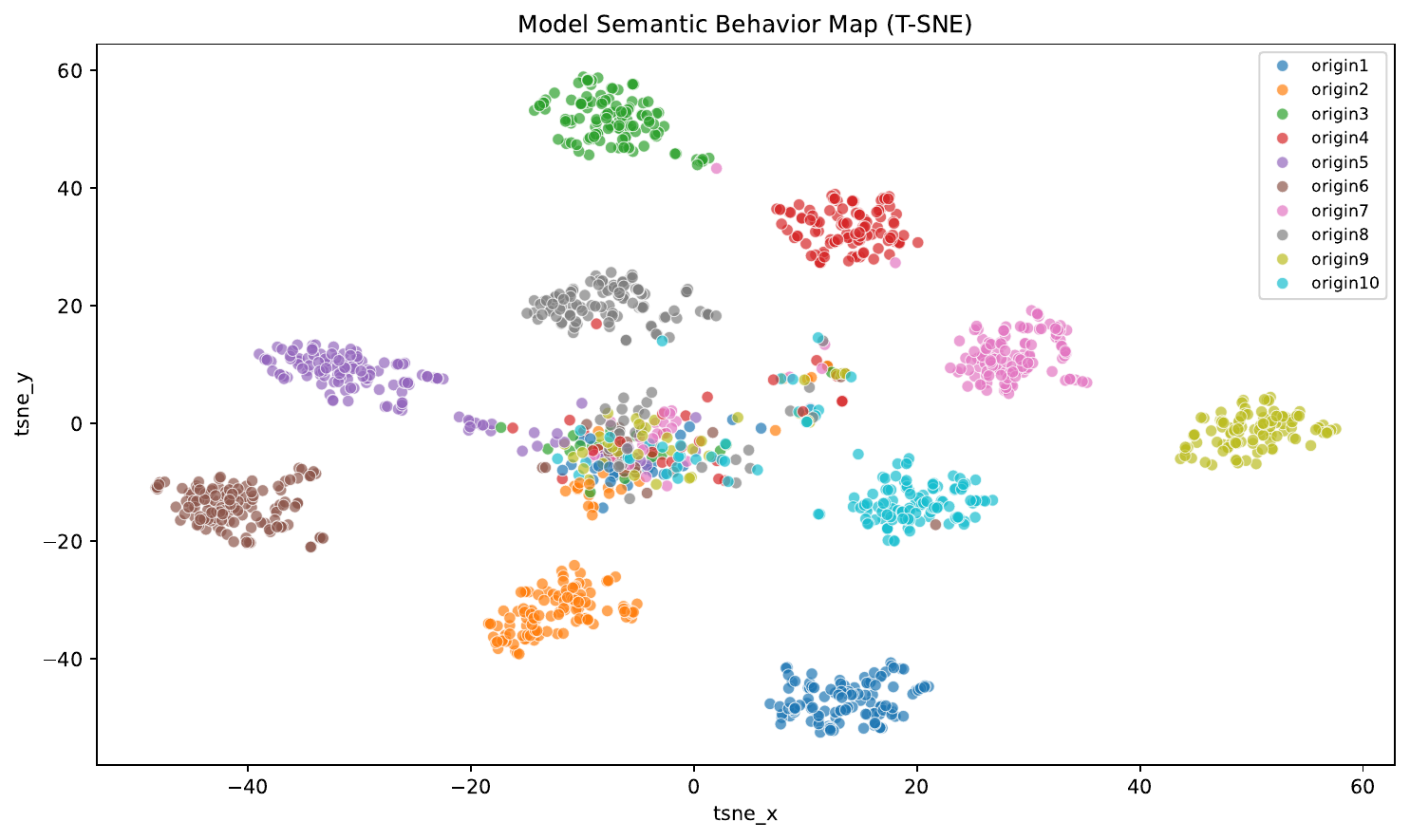}
        \caption{Origin-space t-SNE (10 tasks, 5 new models) - 2}
        \label{fig:tsne-10p-2b}
    \end{minipage}
    \label{fig:tsne-dispersion_2}

\end{figure*}

\begin{figure*}[htbp]
\centering

\begin{minipage}[t]{0.31\textwidth}
  \includegraphics[width=\linewidth]{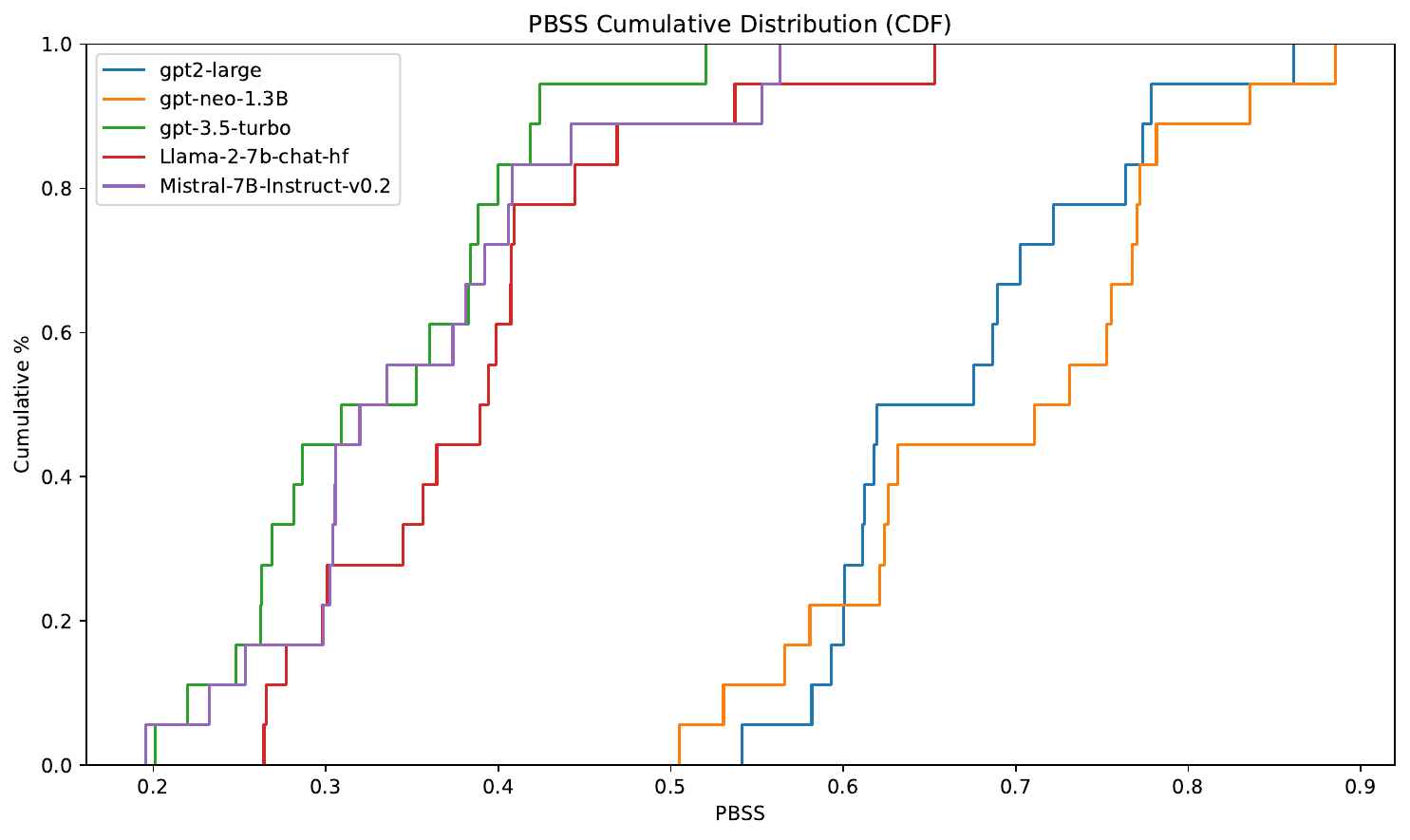}
  \captionof{figure}{PBSS CDF for MiniLM-L6: 3 tasks.}
  \label{9-1}
\end{minipage}
\hfill
\begin{minipage}[t]{0.31\textwidth}
  \includegraphics[width=\linewidth]{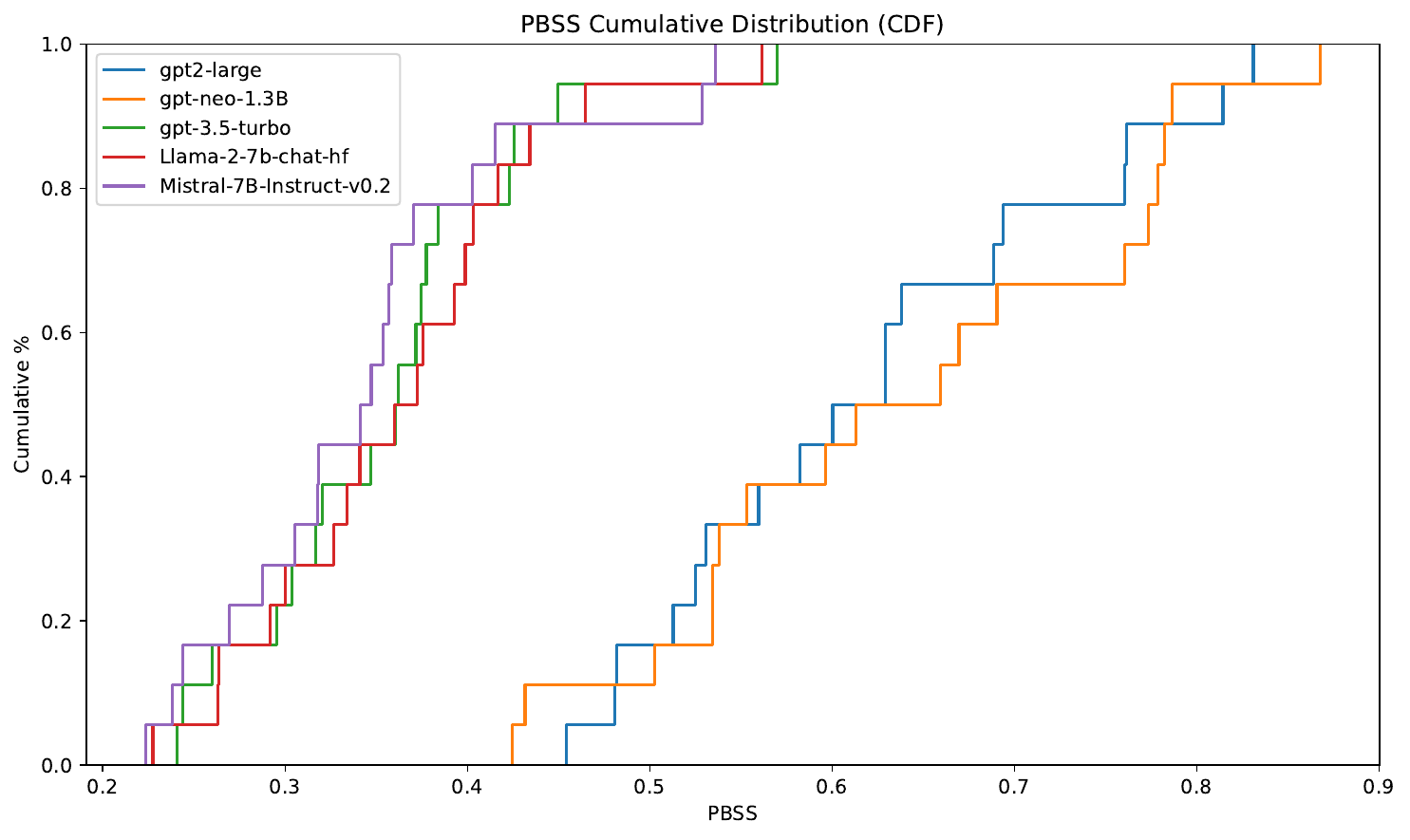}
  \captionof{figure}{PBSS CDF for MiniLM-L12: 3 tasks.}
\end{minipage}
\hfill
\begin{minipage}[t]{0.31\textwidth}
  \includegraphics[width=\linewidth]{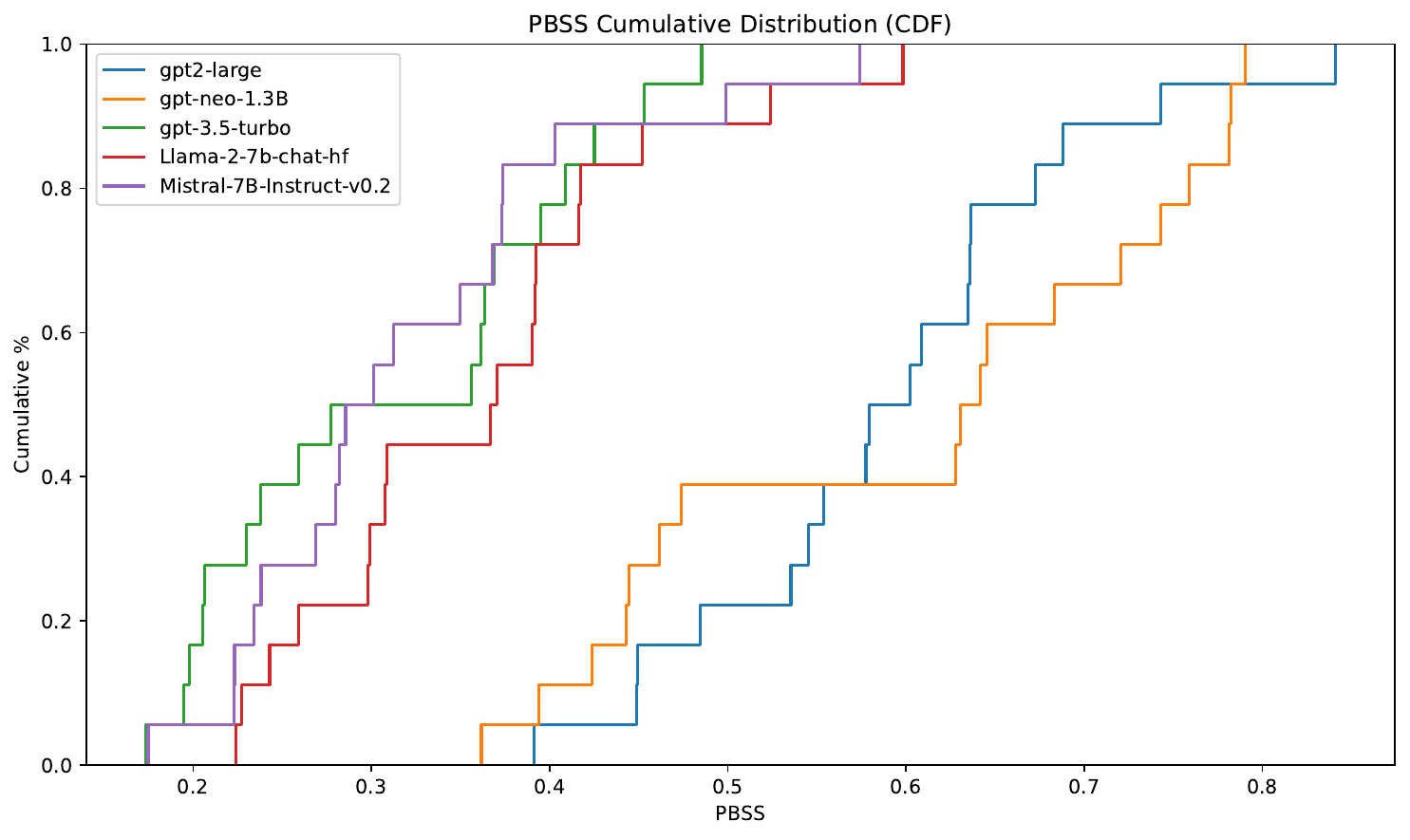}
  \captionof{figure}{PBSS CDF for MPNet: 3 tasks.}
\end{minipage}

\vspace{1em}

\begin{minipage}[t]{0.31\textwidth}
  \includegraphics[width=\linewidth]{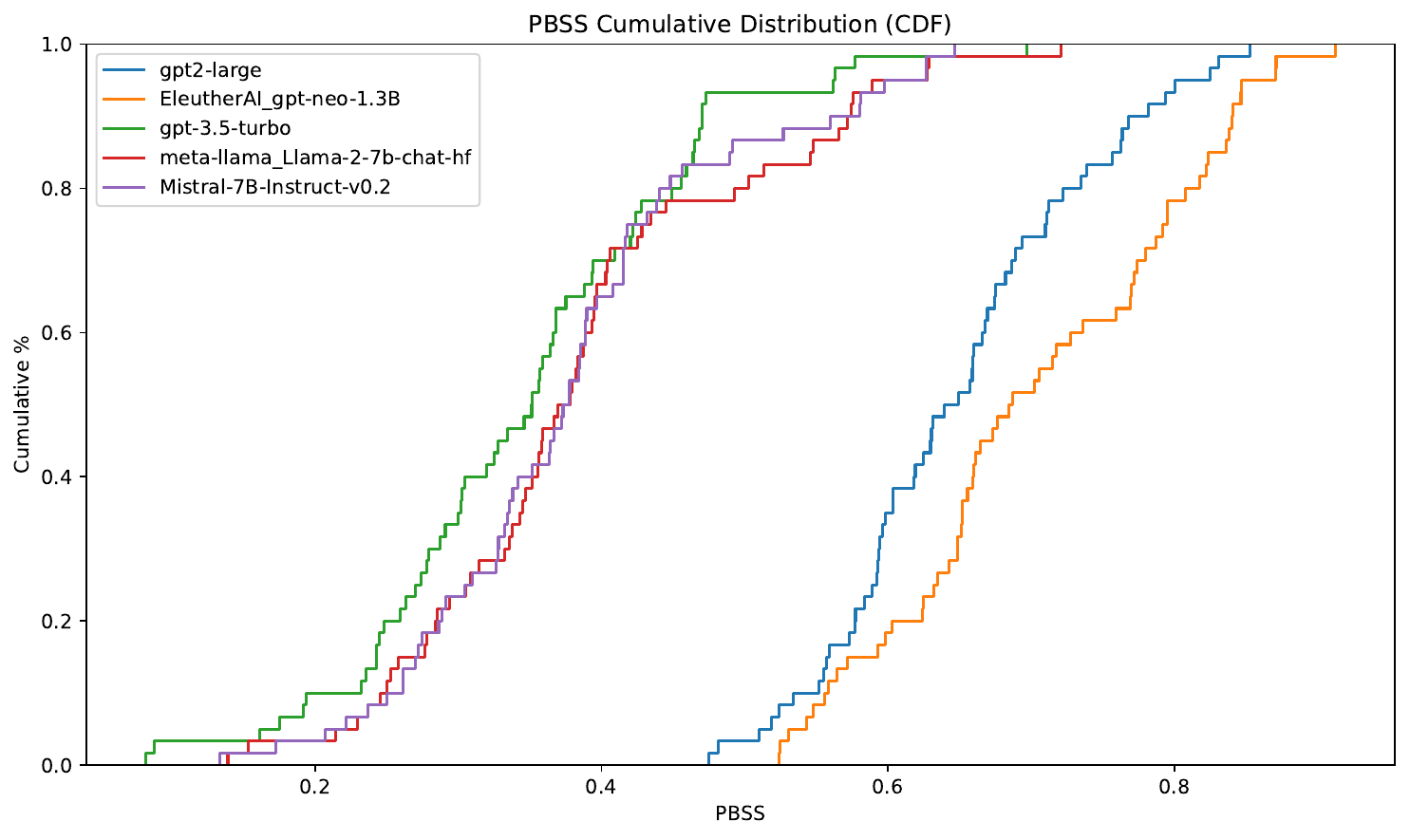}
  \captionof{figure}{PBSS CDF for MiniLM-L6: 10 tasks.}
\end{minipage}
\hfill
\begin{minipage}[t]{0.31\textwidth}
  \includegraphics[width=\linewidth]{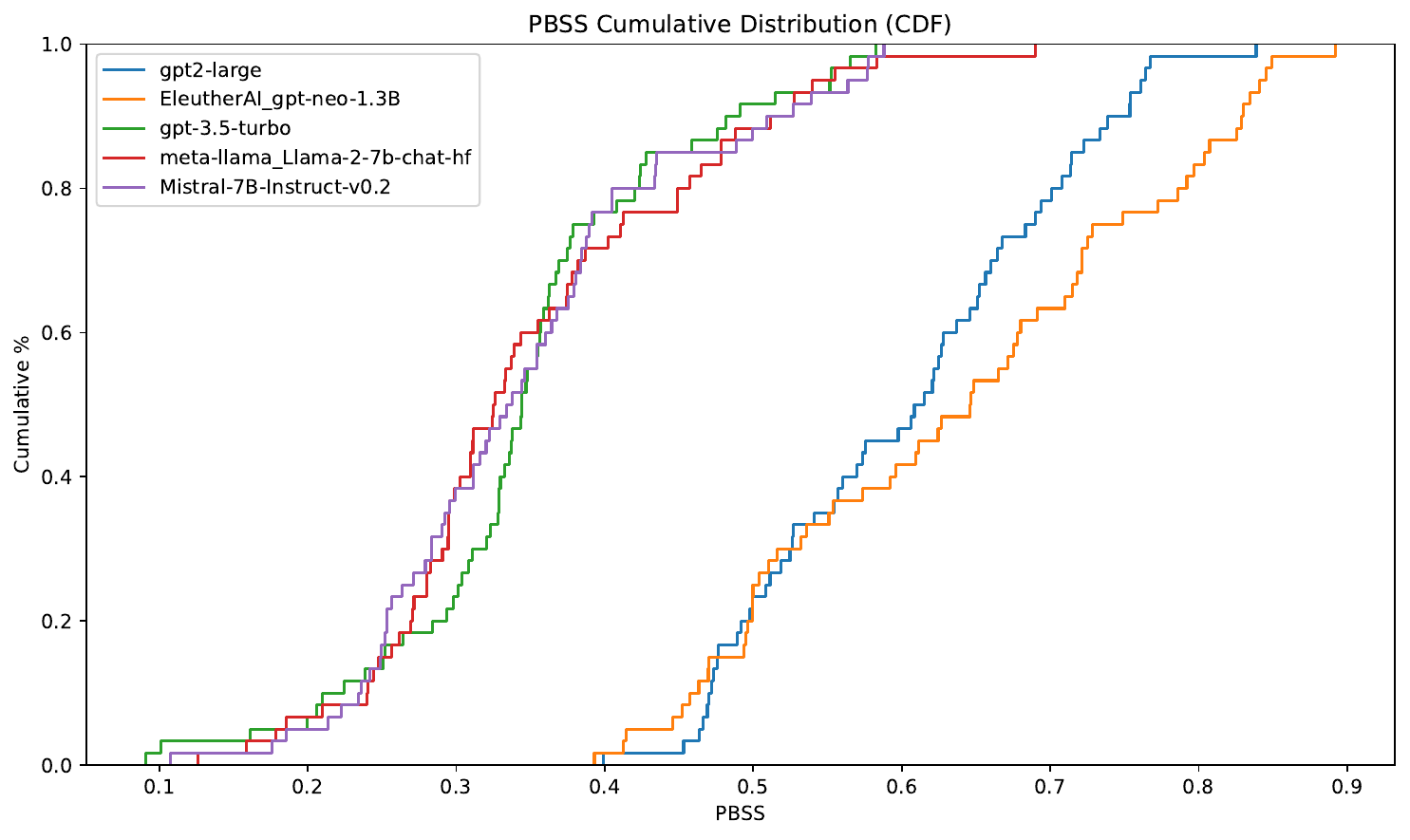}
  \captionof{figure}{PBSS CDF for MiniLM-L12: 10 tasks.}
\end{minipage}
\hfill
\begin{minipage}[t]{0.31\textwidth}
  \includegraphics[width=\linewidth]{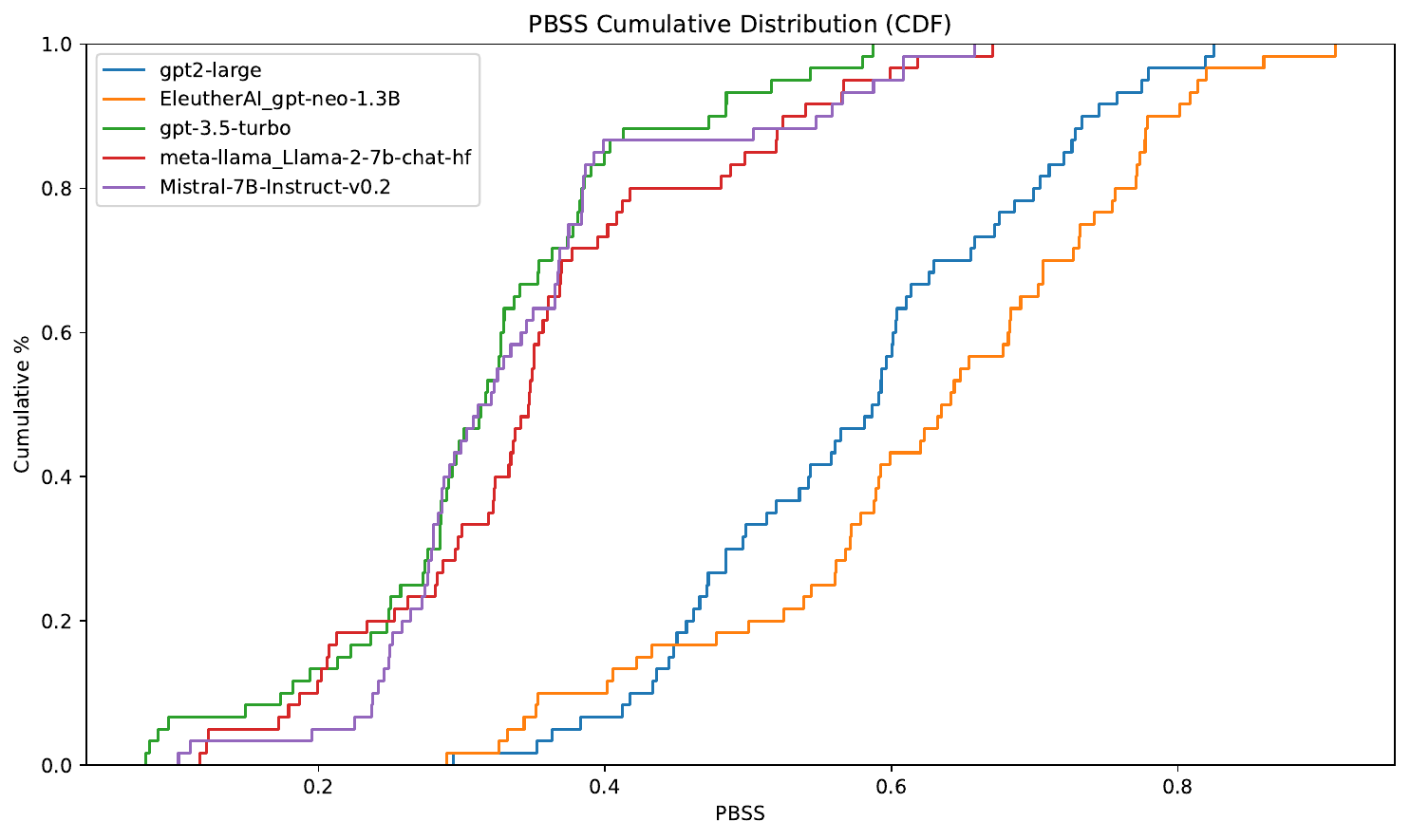}
  \captionof{figure}{PBSS CDF for MPNet: 10 tasks.}
\end{minipage}

\vspace{1em}

\begin{minipage}[t]{0.31\textwidth}
  \includegraphics[width=\linewidth]{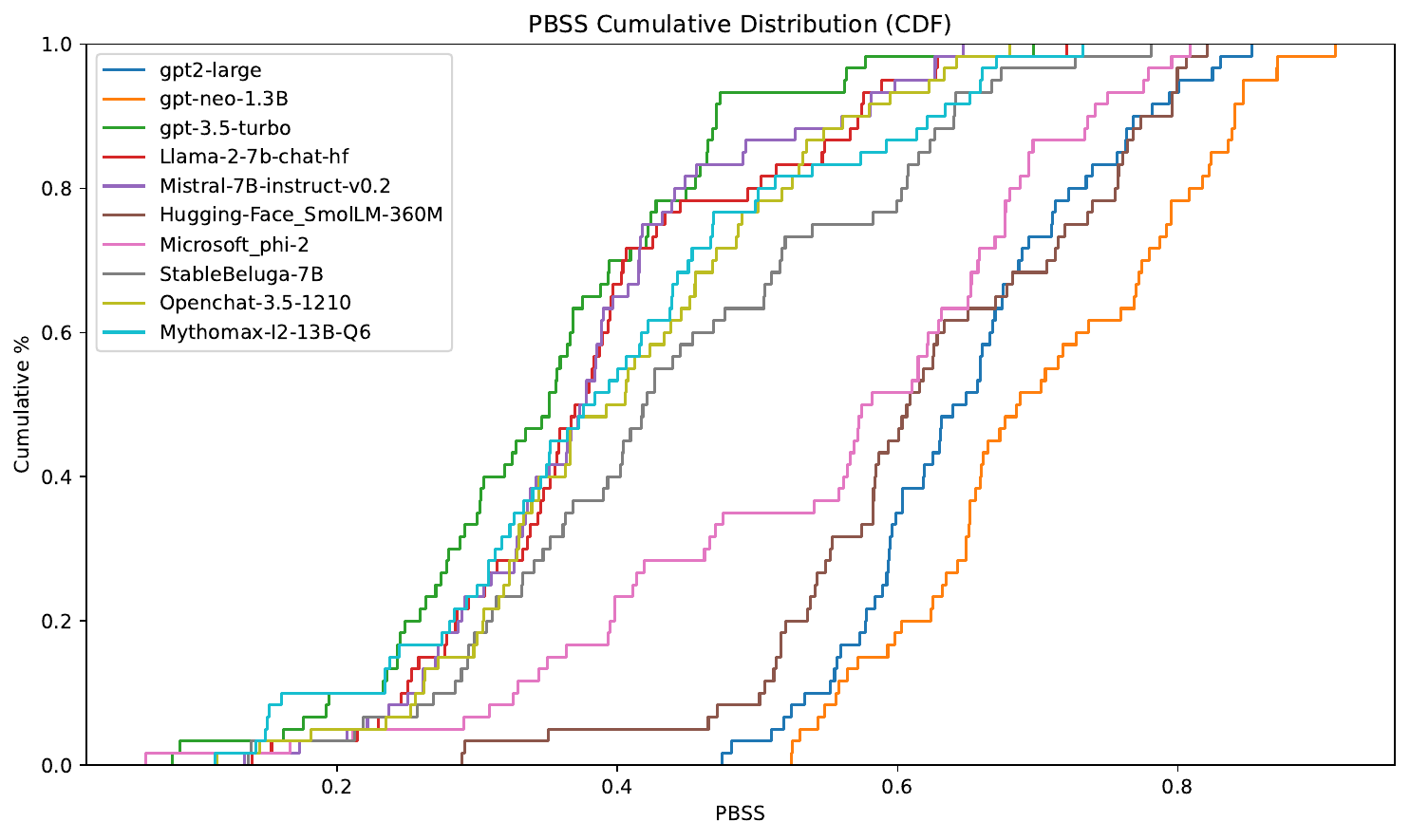}
  \captionof{figure}{PBSS CDF for MiniLM-L6 (alt case).}
\end{minipage}
\hfill
\begin{minipage}[t]{0.31\textwidth}
  \includegraphics[width=\linewidth]{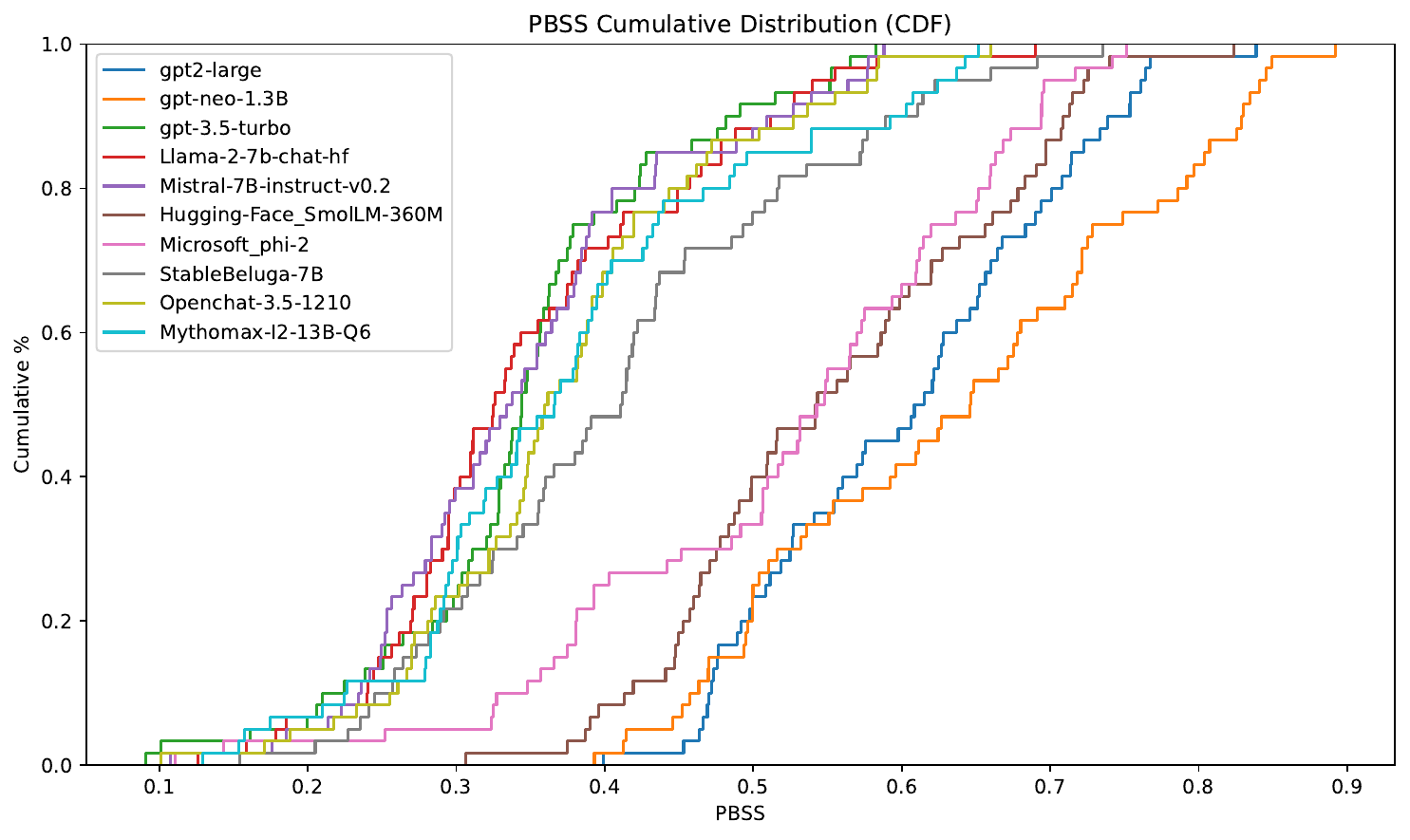}
  \captionof{figure}{PBSS CDF for MiniLM-L12 (alt case).}
\end{minipage}
\hfill
\begin{minipage}[t]{0.31\textwidth}
  \includegraphics[width=\linewidth]{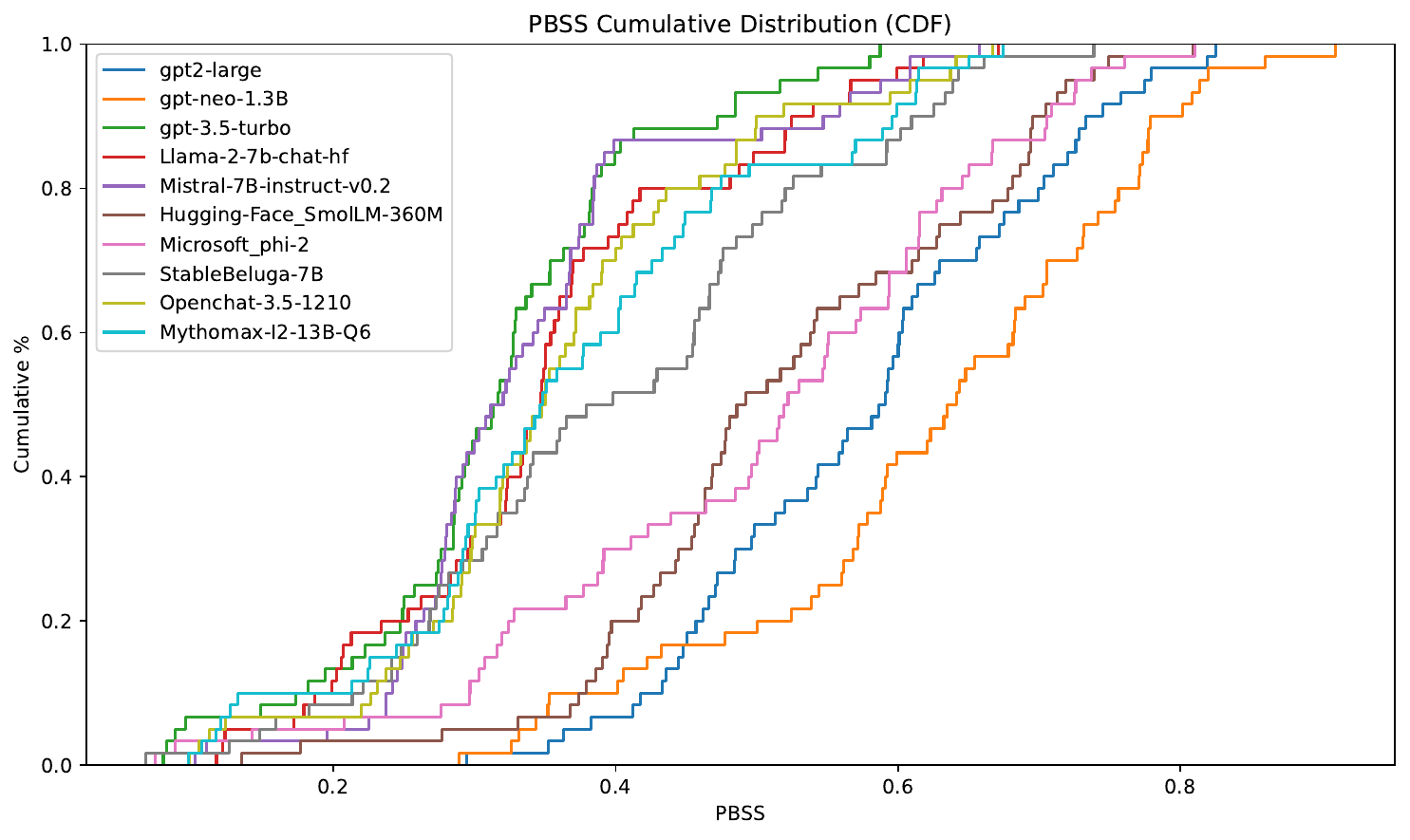}
  \captionof{figure}{PBSS CDF for MPNet (alt case).}
  \label{9-9}
\end{minipage}

\vspace{1em}

\begin{minipage}[t]{0.47\textwidth}
  \includegraphics[width=\linewidth]{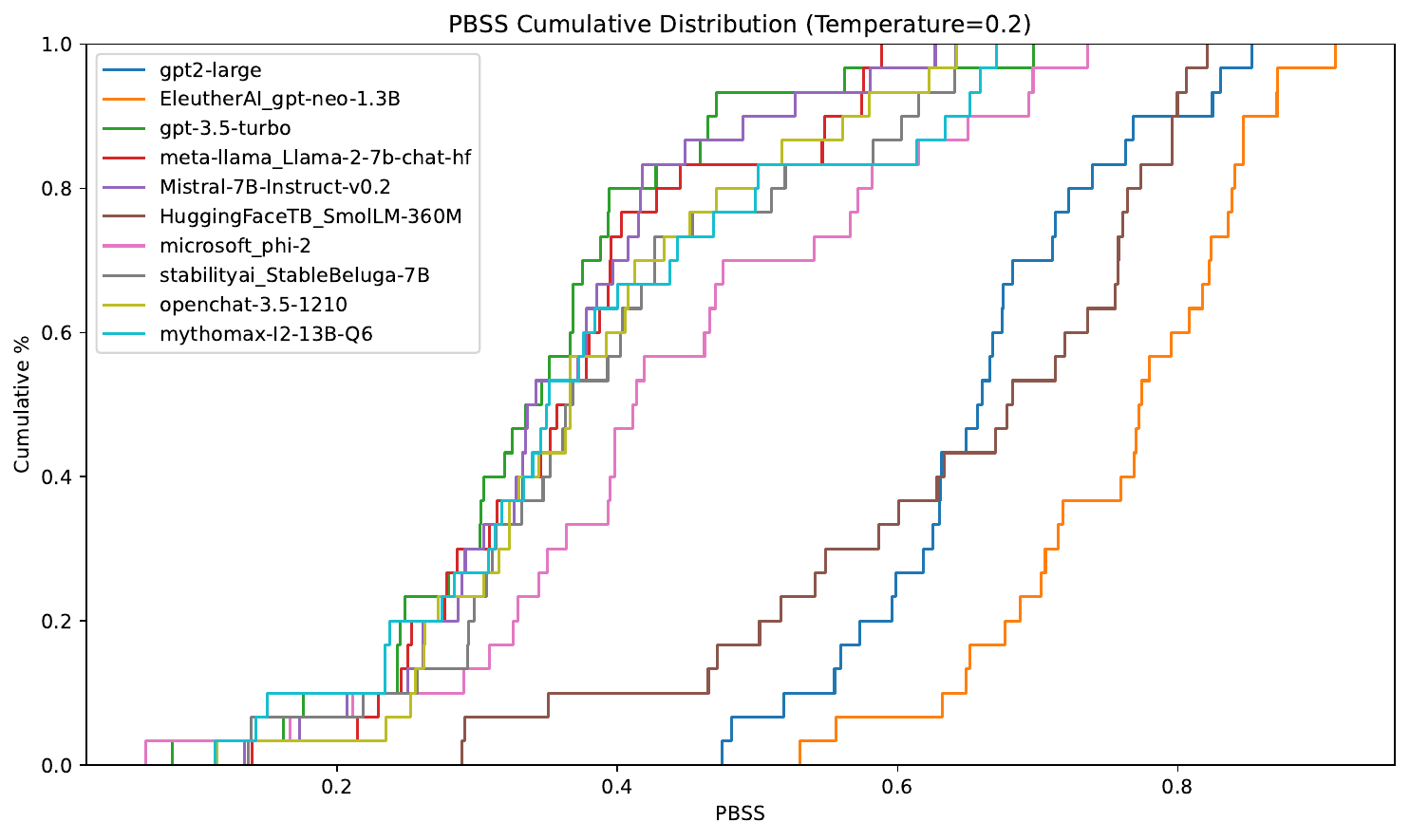}
  \captionof{figure}{PBSS CDF for MiniLM-L6, $T=0.2$.}
\end{minipage}
\hfill
\begin{minipage}[t]{0.47\textwidth}
  \includegraphics[width=\linewidth]{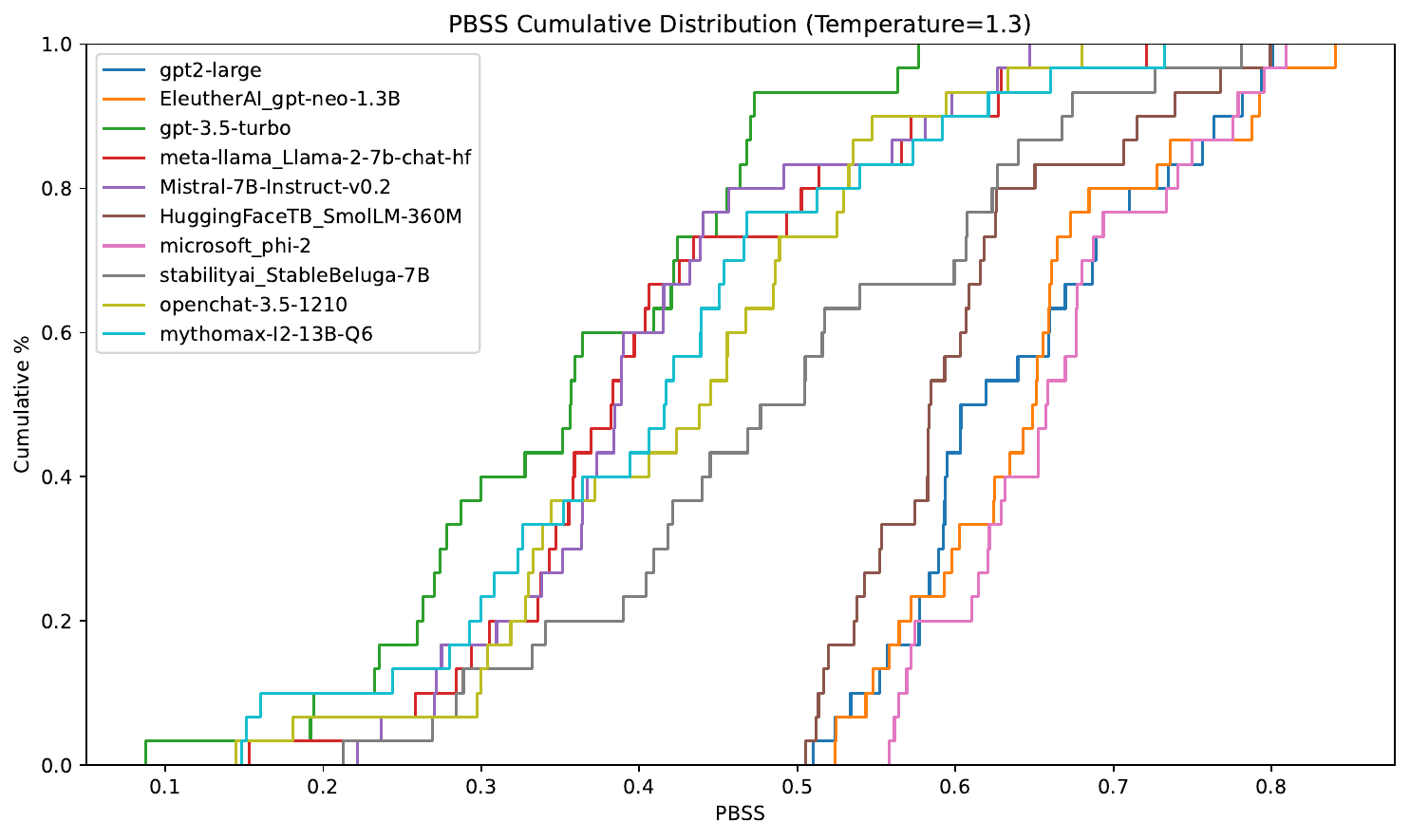}
  \captionof{figure}{PBSS CDF for MiniLM-L6, $T=1.3$.}
\end{minipage}

\end{figure*}

\begin{figure*}[t]
    \centering

    \begin{minipage}[t]{0.31\textwidth}
        \centering
        \includegraphics[width=\linewidth]{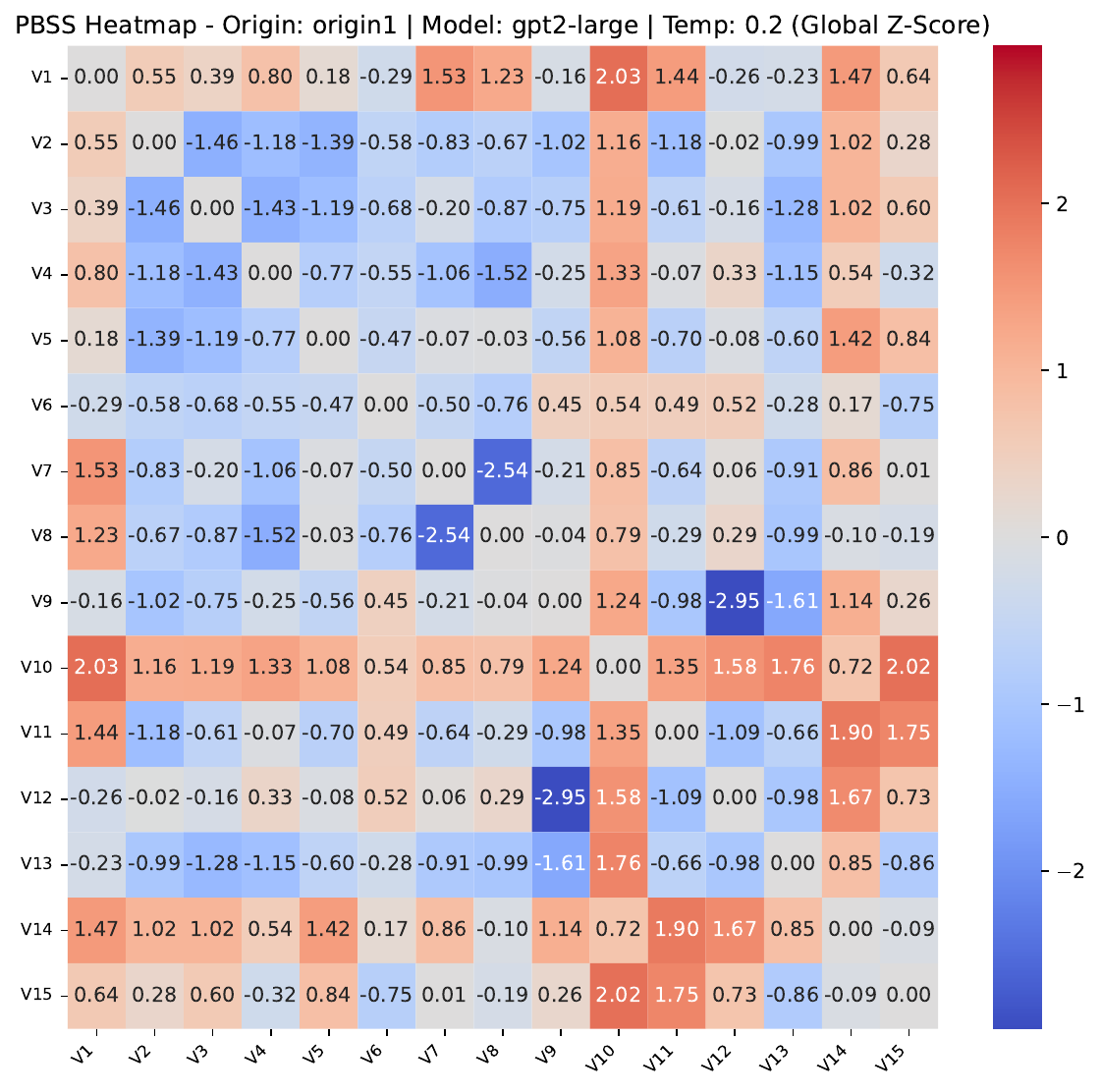}
        \caption{\textbf{GPT-2:} Global Z-Score}
    \end{minipage}
    \hfill
    \begin{minipage}[t]{0.31\textwidth}
        \centering
        \includegraphics[width=\linewidth]{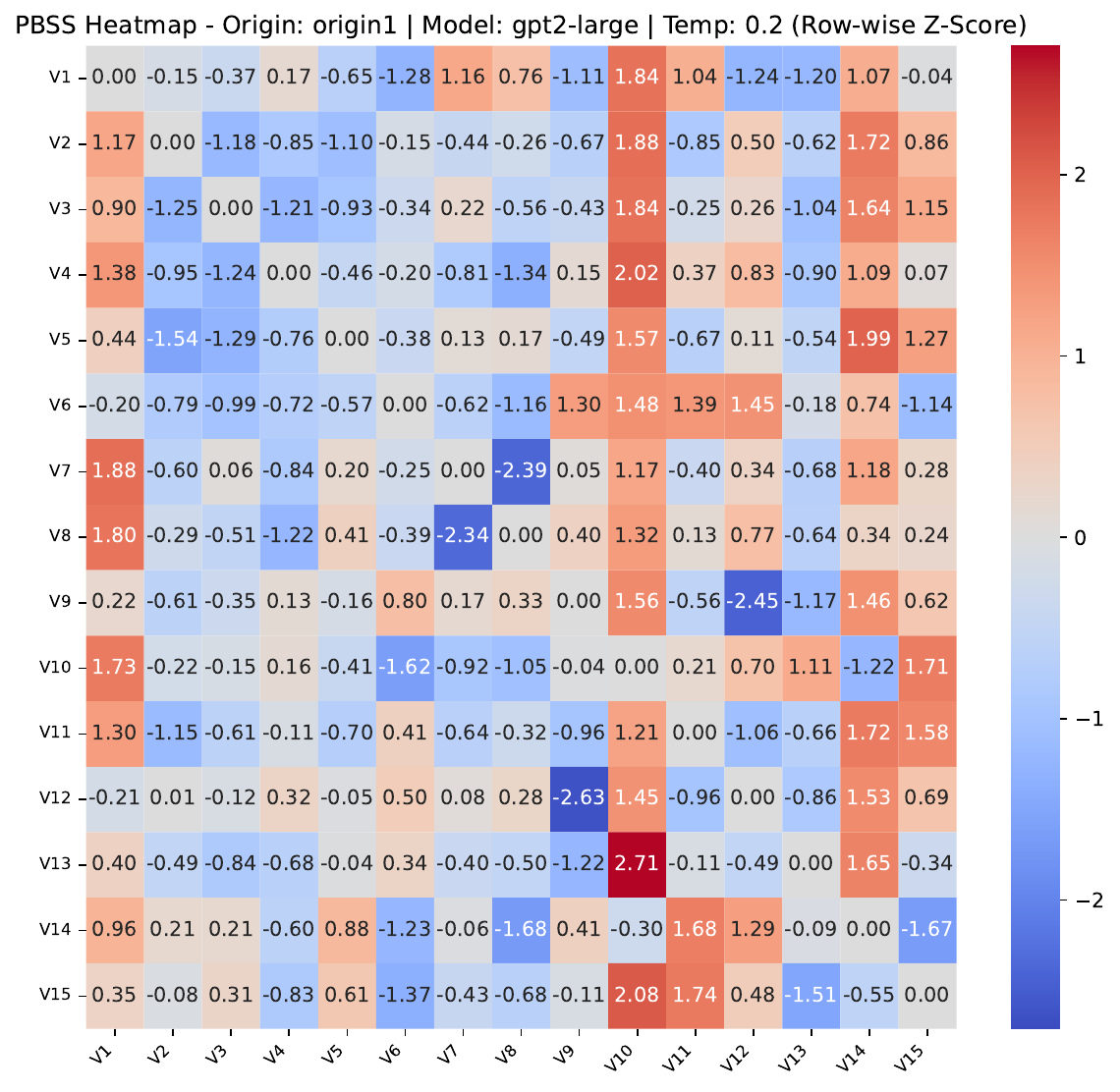}
        \caption{\textbf{GPT-2:} Row-wise Z-Score}
    \end{minipage}
    \hfill
    \begin{minipage}[t]{0.31\textwidth}
        \centering
        \includegraphics[width=\linewidth]{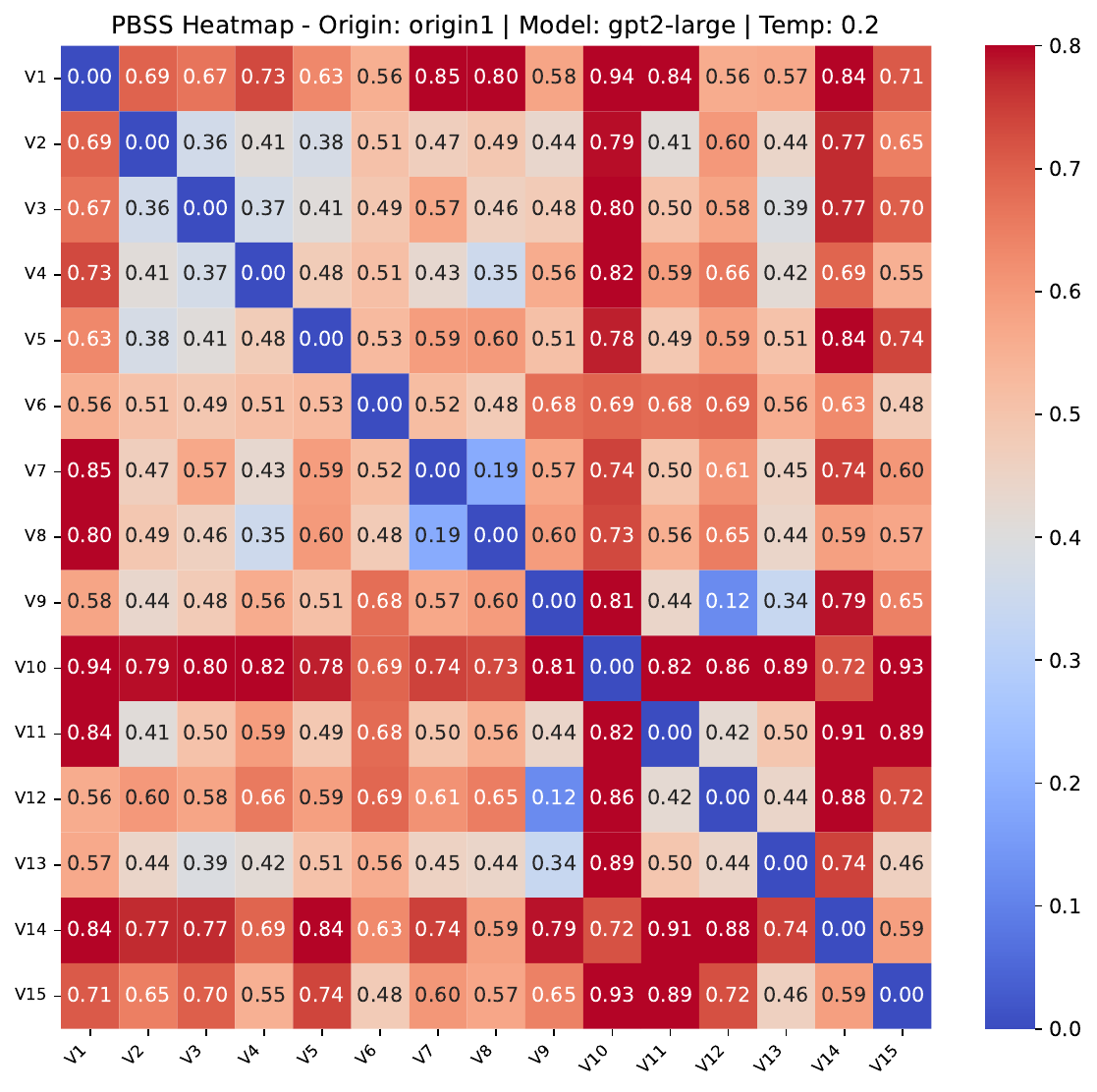}
        \caption{\textbf{GPT-2:} Raw Similarity}
    \end{minipage}

    \vspace{1em}

    \begin{minipage}[t]{0.31\textwidth}
        \centering
        \includegraphics[width=\linewidth]{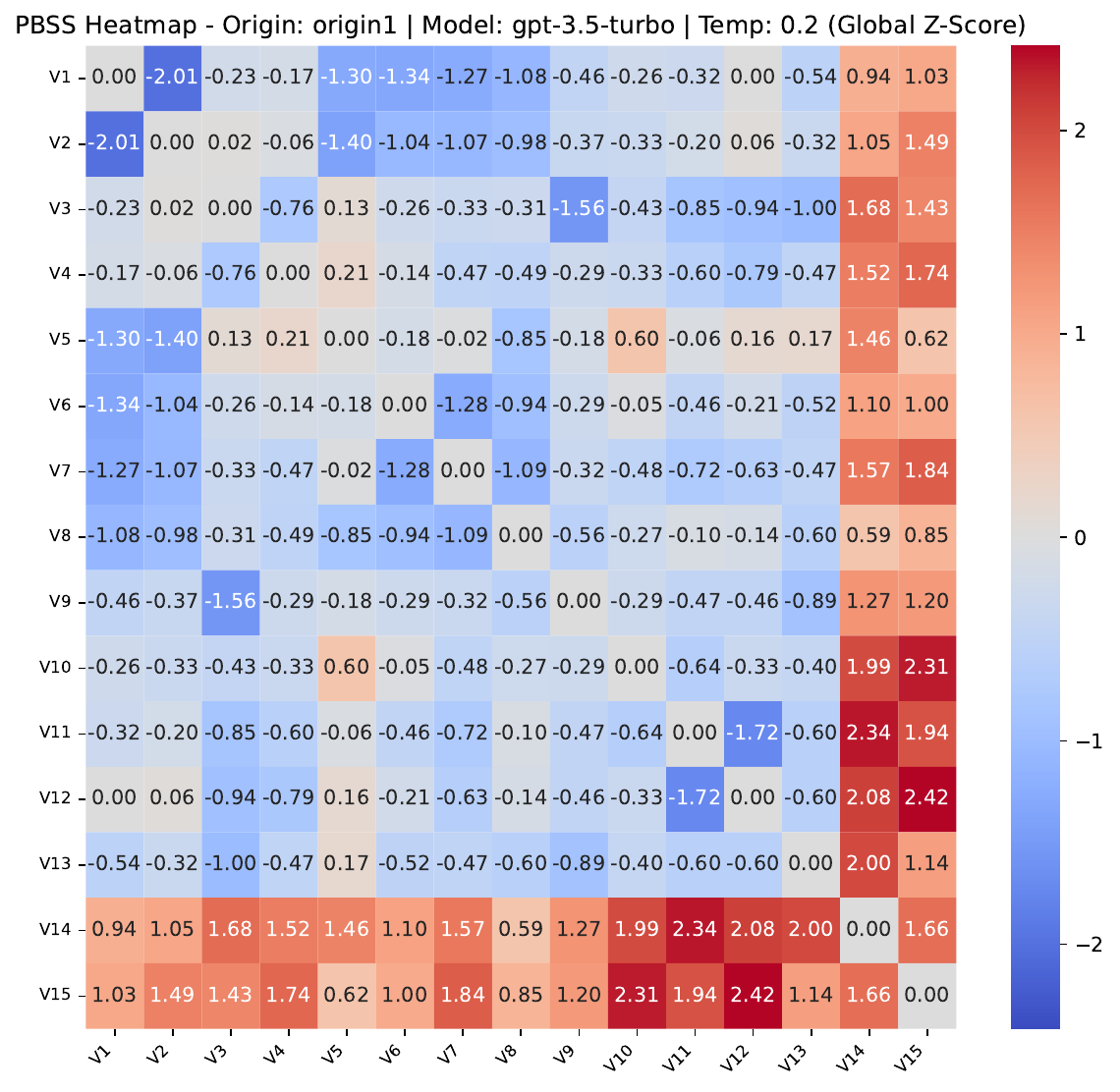}
        \caption{\textbf{GPT-3.5:} Global Z-Score}
    \end{minipage}
    \hfill
    \begin{minipage}[t]{0.31\textwidth}
        \centering
        \includegraphics[width=\linewidth]{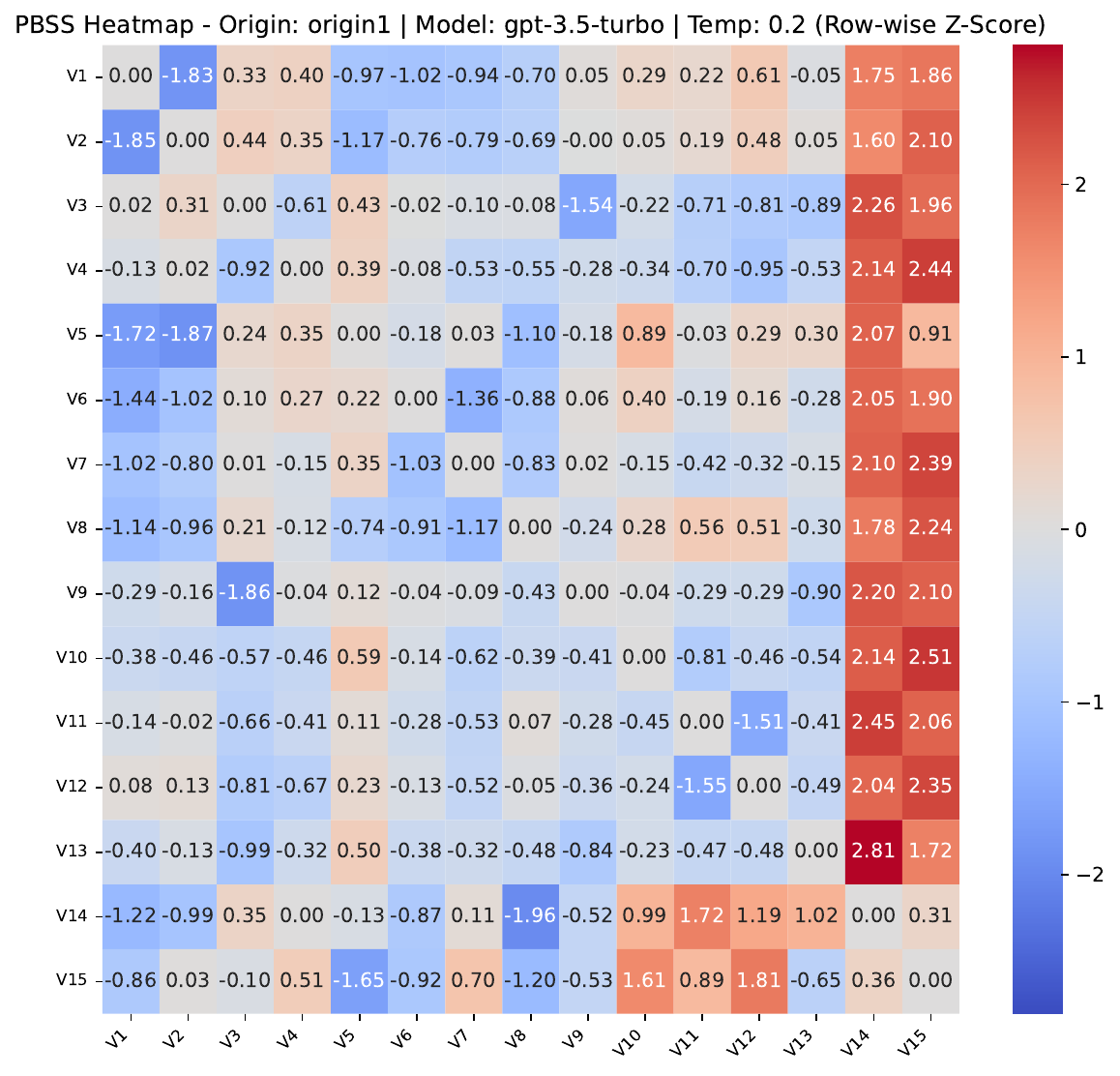}
        \caption{\textbf{GPT-3.5:} Row-wise Z-Score}
    \end{minipage}
    \hfill
    \begin{minipage}[t]{0.31\textwidth}
        \centering
        \includegraphics[width=\linewidth]{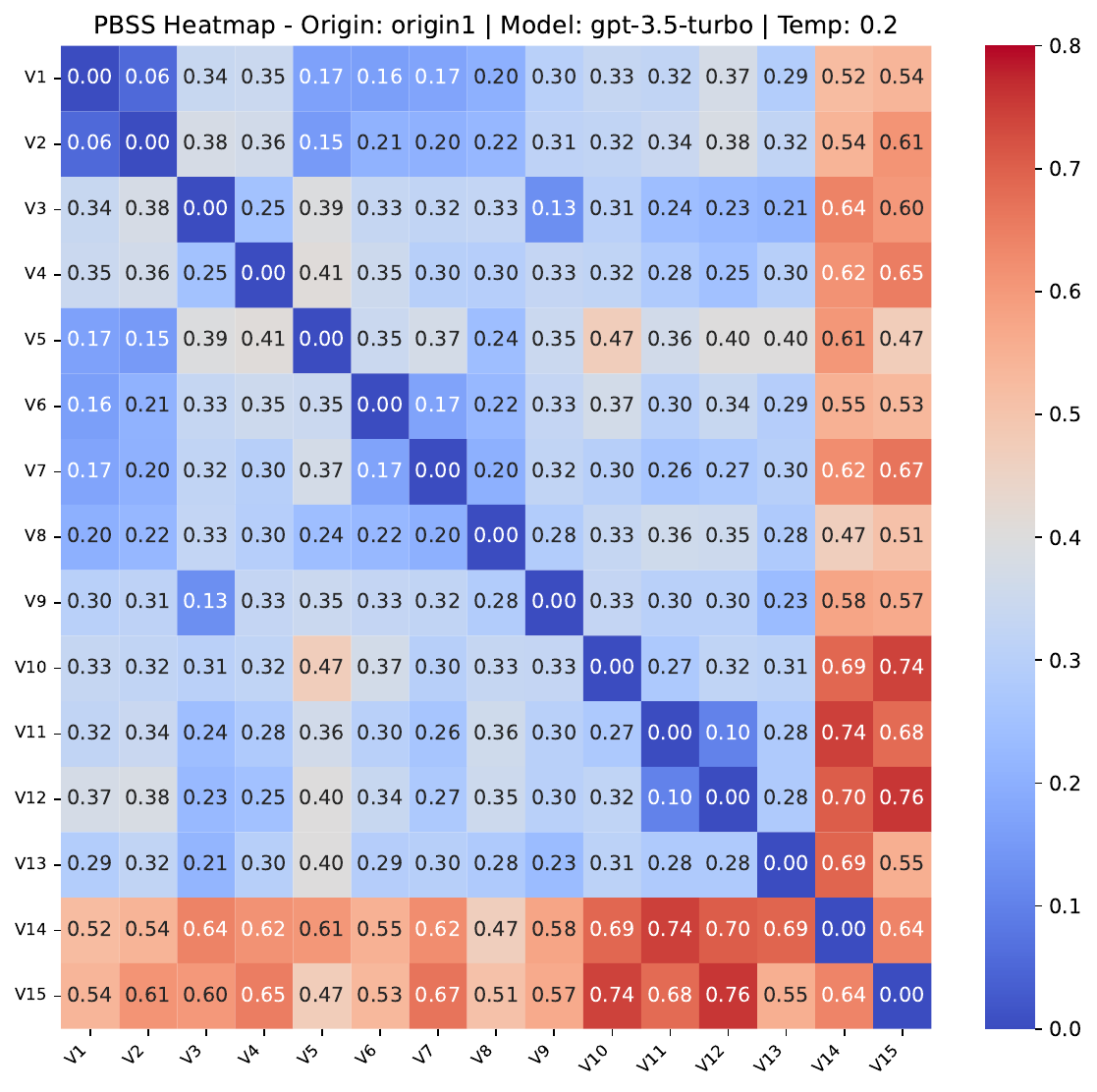}
        \caption{\textbf{GPT-3.5:} Raw Similarity}
    \end{minipage}

    \vspace{1em}

    \begin{minipage}[t]{0.31\textwidth}
        \centering
        \includegraphics[width=\linewidth]{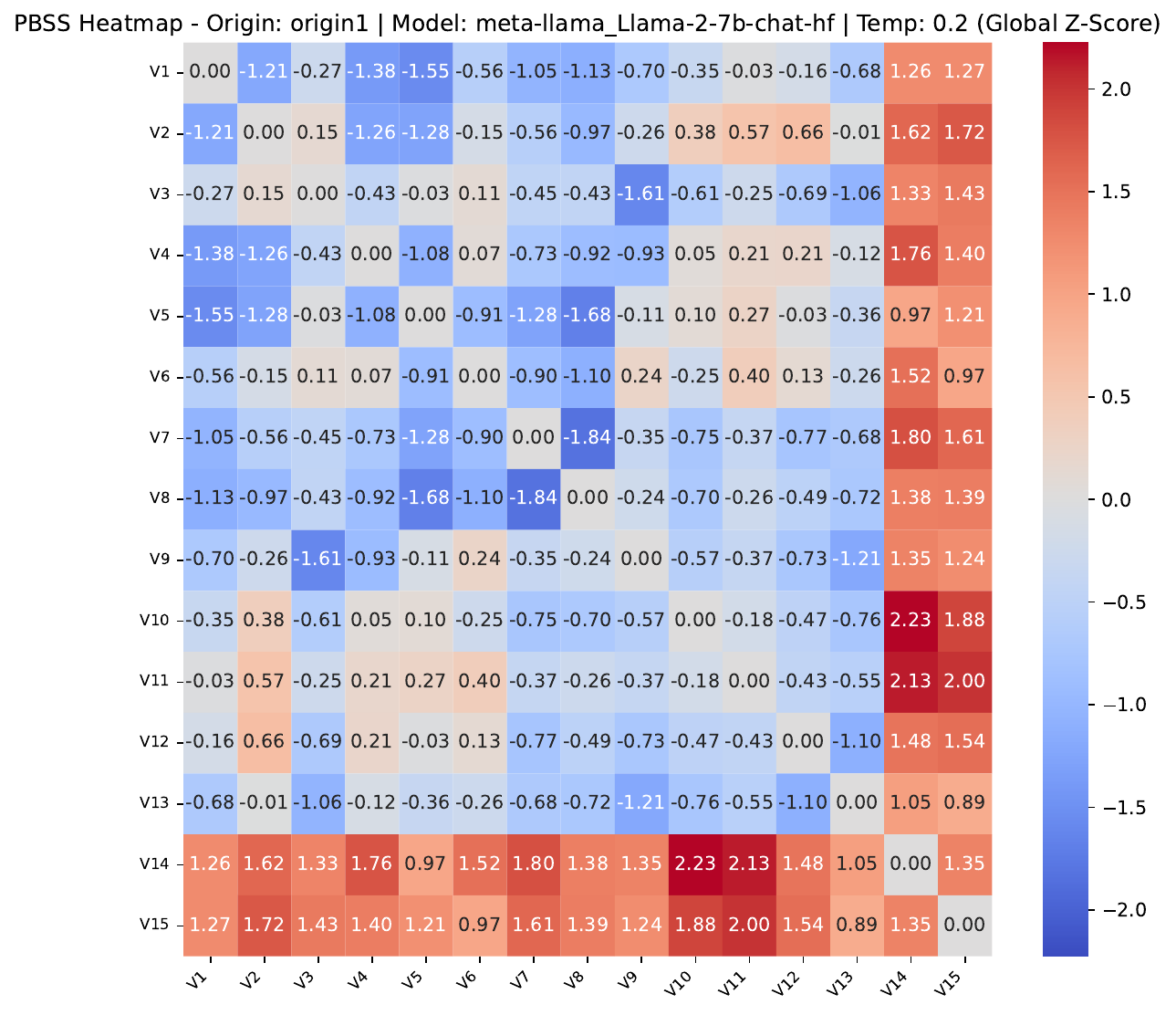}
        \caption{\textbf{LLaMA-2:} Global Z-Score}
    \end{minipage}
    \hfill
    \begin{minipage}[t]{0.31\textwidth}
        \centering
        \includegraphics[width=\linewidth]{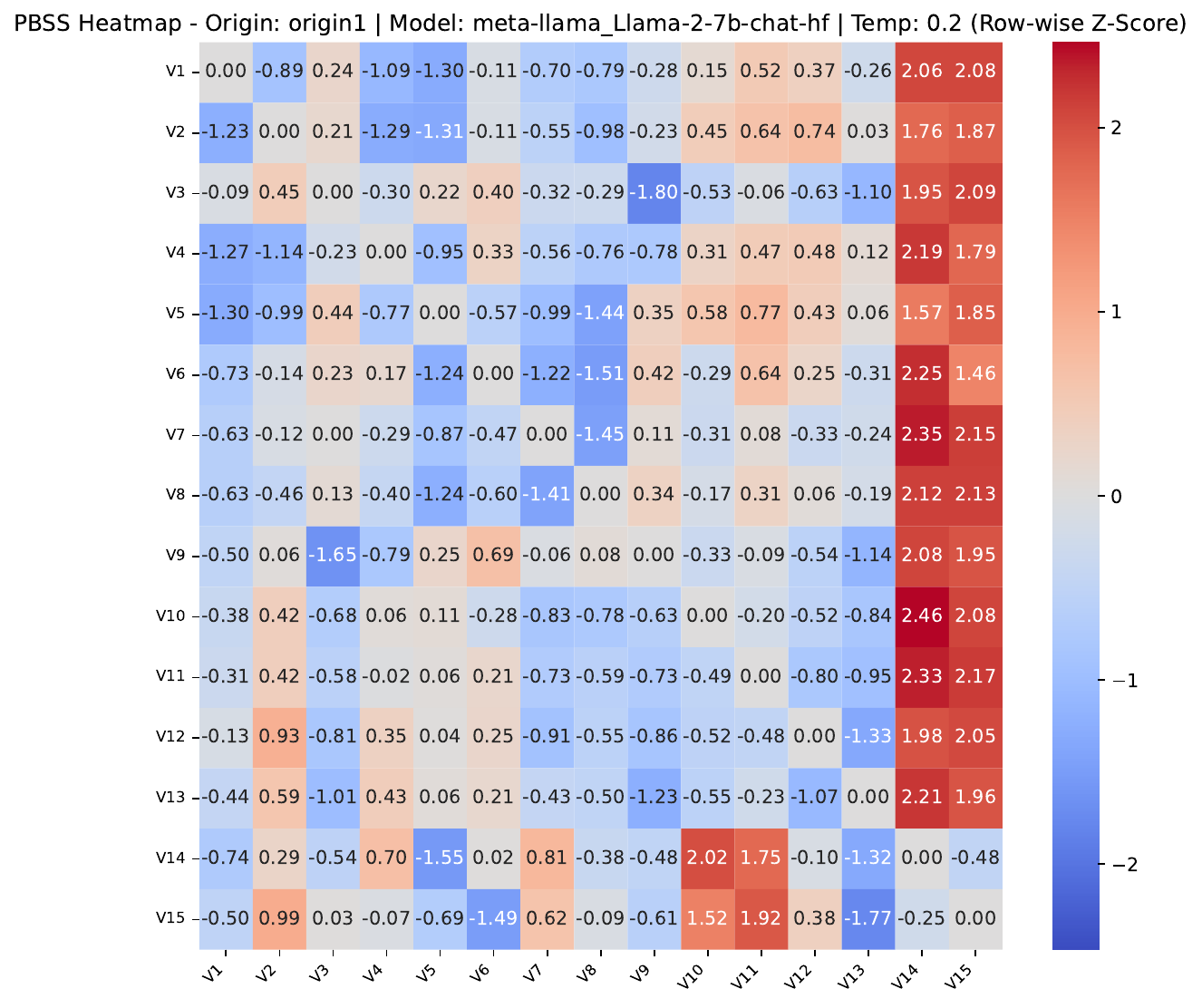}
        \caption{\textbf{LLaMA-2:} Row-wise Z-Score}
    \end{minipage}
    \hfill
    \begin{minipage}[t]{0.31\textwidth}
        \centering
        \includegraphics[width=\linewidth]{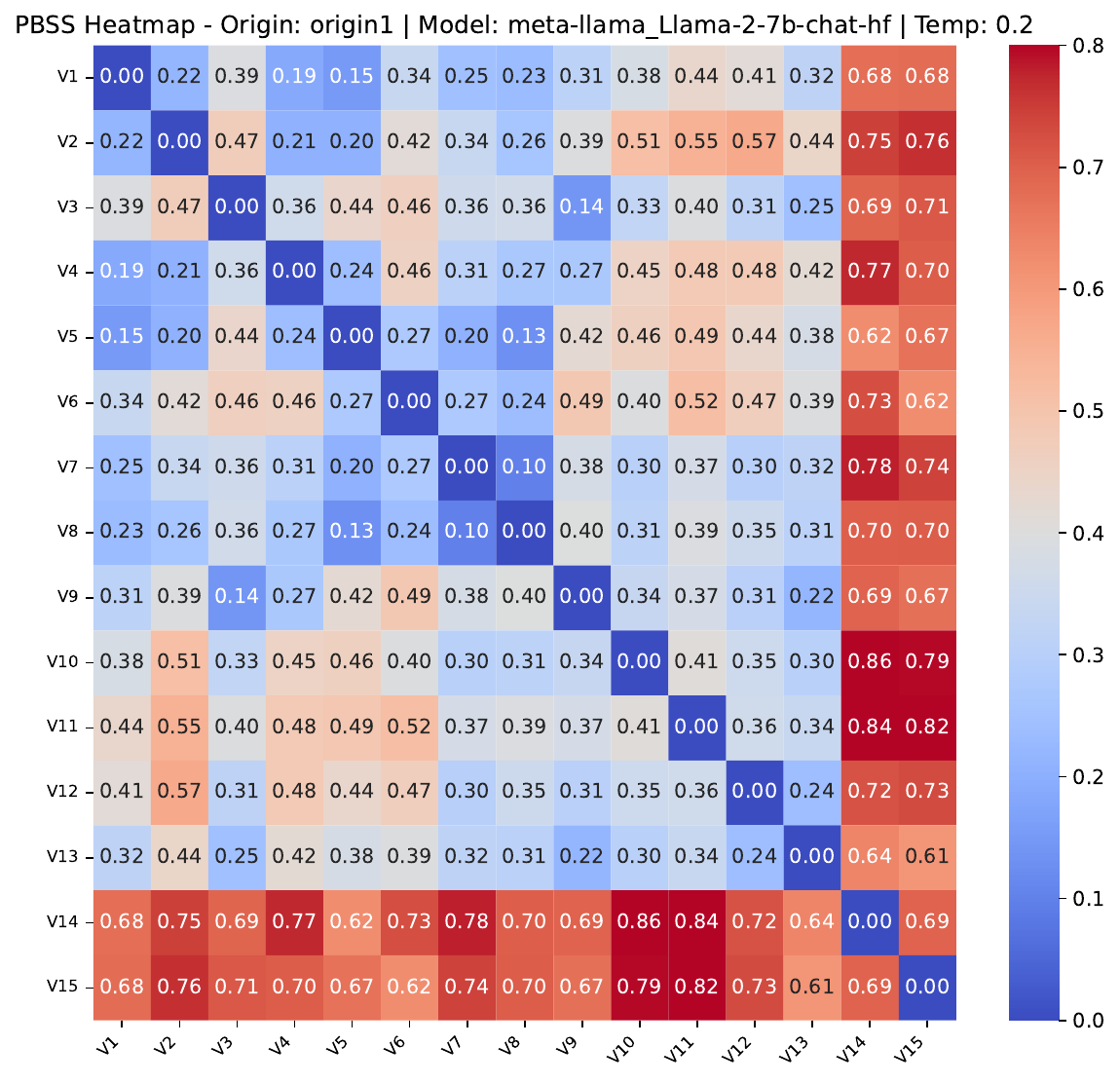}
        \caption*{\textbf{LLaMA-2:} Raw Similarity}
    \end{minipage}

\end{figure*}
\end{document}